\documentclass[12pt]{article}
\usepackage{microtype}
\usepackage{url}
\usepackage{amsmath,amssymb,amsfonts}
\usepackage{algorithmic}
\usepackage{graphicx}
\usepackage{textcomp}
\usepackage{xcolor}
\usepackage{booktabs}
\usepackage{multirow}
\usepackage{tikz}
\usepackage[normalem]{ulem}
\usepackage{float}
\usepackage{subfig}

\usepackage{natbib}
\makeatletter
\newcommand*{\compress}{\@minipagetrue}
\makeatother

\usepackage[margin=1in]{geometry}
\usepackage{setspace}
\usepackage{enumitem}

\DeclareMathOperator*{\argmax}{arg\,max}	
\newcommand{\indep}{\perp \!\!\! \perp}

\providecommand{\keywords}[1]
{
  \small	
  \textbf{\textit{Keywords---}} #1
}


\begin{document}

\title{Bridging the Gap: Towards an Expanded Toolkit for AI-Driven Decision-Making in the Public Sector}

\author{
    Unai Fischer-Abaigar\textsuperscript{1,2,}\thanks{Corresponding author.} 
    \and
    Christoph Kern\textsuperscript{1,2,3}
    \and
    Noam Barda\textsuperscript{4}
    \and
    Frauke Kreuter\textsuperscript{1,2,3}
}

\date{}

\maketitle

\begin{center}

\textsuperscript{1}Dept. of Statistics, LMU Munich, Germany \\
\textsuperscript{2} Munich Center for Machine Learning, Germany \\
\textsuperscript{3} Joint Program in Survey Methodology, University of Maryland, USA \\
\textsuperscript{4} Dept. of Software and Information Systems Engineering and Dept. of Epidemiology, Biostatistics and Community Health Sciences, Ben-Gurion University of the Negev, Israel \\
\end{center}

\begin{abstract}
AI-driven decision-making systems are becoming instrumental in the public sector, with applications spanning areas like criminal justice, social welfare, financial fraud detection, and public health. While these systems offer great potential benefits to institutional decision-making processes, such as improved efficiency and reliability, these systems face the challenge of aligning machine learning (ML) models with the complex realities of public sector decision-making. In this paper, we examine five key challenges where misalignment can occur, including distribution shifts, label bias, the influence of past decision-making on the data side, as well as competing objectives and human-in-the-loop on the model output side. Our findings suggest that standard ML methods often rely on assumptions that do not fully account for these complexities, potentially leading to unreliable and harmful predictions. To address this, we propose a shift in modeling efforts from focusing solely on predictive accuracy to improving decision-making outcomes. We offer guidance for selecting appropriate modeling frameworks, including counterfactual prediction and policy learning, by considering how the model estimand connects to the decision-maker's utility. Additionally, we outline technical methods that address specific challenges within each modeling approach. Finally, we argue for the importance of external input from domain experts and stakeholders to ensure that model assumptions and design choices align with real-world policy objectives, taking a step towards harmonizing AI and public sector objectives.
\end{abstract}

\vspace{1em} 

\noindent
\newpage
\textbf{Highlights:}
\begin{itemize}[noitemsep, topsep=0pt] 
     \item Machine learning (ML) is frequently used to support decision-making in the public sector
    \item A key challenge is the misalignment between ML models and the realities of public sector decision-making
    \item We analyze five challenges to investigate how misaligned technical assumptions can lead to erroneous decision-making
    \item We argue for a shift from focusing solely on predictive accuracy to improving decision-making outcomes
    \item We offer guidance on selecting the right modeling framework, with a focus on causal machine learning and stakeholder input
\end{itemize}

\vspace{1em} 

\keywords{automated decision-making, reliable artificial intelligence, public policy, causal machine learning}
\doublespacing
\section{Introduction}
Automated decision-making (ADM) systems are increasingly being adopted across the public sector \citep{alfterAutomatingSocietyReport2020a, mitchellAlgorithmicFairnessChoices2021, levyAlgorithmsDecisionMakingPublic2021}, often relying on AI models to address a wide array of problem domains, including critical areas such as predictive policing \citep{lumPredictServe2016}, criminal justice \citep{angwin2016machine, mckayPredictingRiskCriminal2020}, fraud detection in government \citep{engstromGovernmentAlgorithmArtificial2020b}, child abuse prevention \citep{chouldechovaCaseStudyAlgorithmassisted2018a}, tax audit selection \citep{blackAlgorithmicFairnessVertical2022},  early warning systems in public schools \citep{perdomoDifficultLessonsSocial2023c}, credit scoring \citep{kozodoiFairnessCreditScoring2022}, profiling of job seekers \citep{desiereUsingArtificialIntelligence2021, kortnerPredictiveAlgorithmsDelivery2021, bachImpactModelingDecisions2023}, development aid \citep{kuzmanovic2024causal} and public health \citep{potashPredictiveModelingPublic2015}. Despite expectations of enhancing decision-making by improving reliability, objectivity, efficiency and uncovering factors that traditional institutional processes may overlook, ADM systems face considerable challenges \citep{barocasFairnessMachineLearning2019, costonValidityPerspectiveEvaluating2023a, engstromGovernmentAlgorithmArtificial2020b, wangPredictiveOptimizationLegitimacy2023, levyAlgorithmsDecisionMakingPublic2021}. Real-world examples demonstrate shortcomings, ranging from racial and gender bias to systems exhibiting poor predictive accuracy leading to flawed decision-making \citep{obermeyerDissectingRacialBias2019a, allhutterAlgorithmicProfilingJob2020, angwin2016machine, dressel2018, mayer2020unintended}. Such unintended consequences are particularly concerning due to their significant impact on individuals' lives and the potential reinforcement of systemic biases. Recent legislation, such as the European Union's AI Act, highlights these concerns by establishing regulations for high-risk AI systems \citep{lauxTrustworthyArtificialIntelligence2023}.

A growing body of literature explores the challenges and potential benefits of employing AI systems to enhance decision-making within the public sector \citep{sunMappingChallengesArtificial2019, zuiderwijkImplicationsUseArtificial2021, penchevaBigDataAI2020, wirtzArtificialIntelligencePublic2019}. Moreover, several reviews examine the adoption of AI in government \citep{levyAlgorithmsDecisionMakingPublic2021}, including US federal institutions \citep{engstromGovernmentAlgorithmArtificial2020b} and the EU public sector \citep{vannoordtArtificialIntelligencePublic2022, AtlasAutomationAutomated}. These reviews cover a wide range of challenges, primarily focusing on institutional, ethical and legal implications of using ADM in the public sector. 

In this article, we focus on challenges that arise from a misalignment between the technical assumptions underlying machine learning (ML) models and the realities of decision-making in complex public sector environments. Specifically, we will discuss AI-driven decision-making used for the allocation of scarce resources in the public sector, where decisions involve determining whether individuals qualify to receive specific interventions or services \citep{kupplerFairPredictionsJust2022b}. Our focus is on ADM systems that do not rely on manually encoded rules, but rather use supervised ML models to learn patterns from historical data to predict relevant outcomes that inform decision-making. Although ML approaches can vary widely, ranging from support vector machines to neural networks, we aim to keep our discussion relevant across different models by exploring the general limitations and challenges of using supervised ML for public sector decision-making. Throughout the text, we use terms like AI, ML and predictive algorithm interchangeably to refer to the computational model underlying the ADM system.

Decision-making in these environments often takes place in dynamic, evolving social contexts, which can conflict with the explicit formalization requirements demanded by ML models  \citep{levyAlgorithmsDecisionMakingPublic2021, mitchellAlgorithmicFairnessChoices2021, amarasingheExplainableMachineLearning2023a, passiProblemFormulationFairness2019}. Technical choices made during model development often rest on implicit assumptions, such as stable data distributions and a straightforward link between prediction and decision-making, that may not hold in these complex settings. For example, policy objectives are often shaped by multiple stakeholders, political compromises and competing goals \citep{levyAlgorithmsDecisionMakingPublic2021, coyleExplainingMachineLearning2020a}, making it difficult to translate them into clearly defined objectives for ML systems. When the assumptions behind the technical model construction do not align with the deployment context, there is a risk of developing systems that fail to capture the complexities of real-world decision-making, potentially leading to adverse outcomes upon model deployment.

Consider, for example, a public employment service (PES) office that aims to determine which job seekers should participate in job programs to increase their re-integration chances. The PES wants to deploy ML to learn the optimal assignment of support to job seekers based on data collected as part of their daily operations. However, the PES now faces two critical sets of interconnected complications: first, while their data may include detailed records of labor market histories, they are operating in a complex and dynamic social environment which raises questions of distribution shift, feedback loops and the challenge of accounting for the effect of competing (current) and previous job support programs. Second, the PES needs to ensure that the predictions can effectively be integrated in their current decision-making practices. This may require model guarantees to build caseworker trust in the predictions and ensuring that other relevant objectives and constraints are sufficiently incorporated in the system. All these issues require careful consideration in the model design choices. A misalignment between technical assumptions and problem setting, such as building a model under the implicit assumption that labor market characteristics remain invariant, may result in unintended consequences, such as an allocation mechanism that might become unreliable over time. 

Efforts to analyze challenges from a technical perspective are ongoing and focus on connecting methodological AI research with the unique demands of high-stakes decision-making. These efforts explore various subdimensions of this complex issue, including training data quality \citep{shahbazi.et.al.2023}, target variable bias \citep{guerdanGroundLessTruth2023a} and uncertainty \citep{gruberSourcesUncertaintyMachine2023a, kaiserUncertaintyawarePredictiveModeling2022a}. Furthermore, active research develops frameworks to examine the conditions under which the usage of predictive algorithms for high-stakes decision-making can be justified \citep{costonValidityPerspectiveEvaluating2023a, wangPredictiveOptimizationLegitimacy2023}.

In this work, we identify and analyze misalignments that commonly occur between ML models and public sector decision-making. Guided by the `ADM process model' \citep{Gerdon2022}, we focus on how models connect with their wider real-world deployment context by examining both the data assumptions (model input) and how models are integrated into the decision-making process (model output). Using the lens of misalignment developed here, we build on the recent technical literature on ML and decision-making to isolate five specific challenges that we consider to exemplify the type of issues that can occur at these two interfaces: distribution shift, label bias and the influence of past decision-making on the input side, and competing objectives and constraints and human-in-the-loop interactions on the output side. We analyze each of these challenges to better understand how misaligned technical assumptions can lead to erroneous decision-making and adverse outcomes for affected individuals in public sector environments.

Through our analysis, we find that standard ML methods often rely on assumptions that do not fully account for the complexities of public sector decision-making. In response, we propose a shift in modeling efforts from focusing solely on predictive accuracy to improving decision-making outcomes. We argue that achieving this shift may, in certain cases, require alternative  modeling techniques that extend purely predictive models, and more directly center on the goal of decision-making. With this in mind, we highlight promising developments in causal machine learning, including counterfactual prediction and policy learning. Within each modeling framework, we summarize technical methods that provide (partial) solutions to the identified challenges. To guide practitioners in selecting the right approach, we clarify the assumptions underlying each framework, specifically addressing, how the model estimand connects to the utility of the decision-maker and the data and assumptions required for reliable estimation.

By examining these frameworks through the lens of public sector decision-making, we want to encourage technical practitioners to carefully consider the assumptions behind different modeling approaches and expand their toolbox to include methods that may be better suited for complex, real-world decision-making. For policymakers, domain experts and other stakeholders, we outline which external input is important to help model developers make the right assumptions to inform model design.

While we do discuss several risks resulting from the assumptions made during model development, zooming out to the institutional and societal context raises more complex issues. For instance, institutional and cultural biases embedded in historical data as well as data collection methods and processing can significantly contribute to discriminatory decision-making \citep{JANSSEN2016371, FOUNTAIN2022101645}. Algorithmic systems may also reinforce existing structural inequalities by formalizing problematic decision-making practices \citep{kolkman2020usefulness} or empowering institutions with unjust goals. The continued digitization of bureaucratic processes, particularly when multiple institutions and systems interact, can create new risks, such as making it harder to correct errors across systems or systematically excluding specific user groups \citep{peeters2018digital}. However, we consider addressing misalignments between assumptions made during model development and the deployment context to be essential for avoiding harmful model design, making it a necessary (though not sufficient) condition for the fair development of AI-driven decision making in the public sector.

This paper is structured as follows. We first explore central (mis)alignment challenges that occur along the ML pipeline when developing and deploying AI systems to support decision-making in the public sector (Section \ref{sec: challenges}). Second, we highlight recent methodological developments that exceed the classical supervised ML paradigm, showing promise in addressing the challenges identified (Section \ref{sec: frameworks}). Third, we discuss the selection of an appropriate modeling approach in a given deployment context (Section \ref{sec: toolkit}). In the discussion, we address broader issues related to ADM in the public sector, specifically highlighting the importance of domain expertise and stakeholder input  (Section \ref{sec: discussion}). Finally, Section \ref{sec:conclusion} provides a concise summary of our findings.

\section{Defining the Gap: Central Challenges in Connecting ML and Decision-Making}
\label{sec: challenges}

Predictive models for ADM systems are designed to inform decisions in (interaction with) dynamic social contexts, which gives rise to a list of fundamental challenges. This includes questions related to choosing adequate model input, as the effectiveness of any ML model is fundamentally linked to the quality of its training data. Ensuring that this data accurately represents the target population is key to avoid biased and unreliable predictions \citep{gruberSourcesUncertaintyMachine2023a}. Securing representative data in the public sector, however, is a complex task. Public sector data is incredibly diverse \citep{dwivediArtificialIntelligenceAI2021, janssenFactorsInfluencingBig2017}, comes in a variety of formats, often lacks structure and encompasses a wide array of data modalities \citep{dwivediArtificialIntelligenceAI2021}. Despite the abundance of data in theory, high-quality data suitable for ML is often not easily available in the public sector \citep{alexopoulosHowMachineLearning2019, sunMappingChallengesArtificial2019}. In many countries, the lack of robust infrastructure to enable data sharing and integration of various data sources can hinder the development of ML models \citep{sunMappingChallengesArtificial2019, wirtzArtificialIntelligencePublic2019}, stemming from issues such as resource constraints, data protection, safety concerns and institutional pushback. These issues are especially concerning for ADM systems, since building a model that is capable of informing future decision-making places significant demands on the training data \citep{hullermeierPrescriptiveMachineLearning2021a, costonValidityPerspectiveEvaluating2023a}.

In addition, it is important to consider how the model output will be integrated into the decision-making process. Rather than improving model performance in isolation, the success of a system should be evaluated based on whether it helps guide decision-making to achieve the intended policy objectives \citep{mitchellAlgorithmicFairnessChoices2021}. Choosing the appropriate modeling setup, requires drawing a connection from broad, often hard to formalize policy goals to the specific target outcomes estimated by the prediction model \citep{levyAlgorithmsDecisionMakingPublic2021}. 

ADM systems aim to identify individuals for targeted interventions, typically with the goal of improving an objective defined as the aggregate of the individual outcomes of interest. For instance, policy makers may seek to improve healthcare in a hospital as a function of individual treatment outcomes or maximize money recovered during tax audits \citep{blackAlgorithmicFairnessVertical2022}. These overarching policy goals may be formalized through an allocation principle that determines the optimal assignment of interventions based on the estimated outcomes \citep{kupplerFairPredictionsJust2022b}. For example, we may choose to intervene when the effect of an intervention is expected to be positive \citep{fernandez-loriaCausalDecisionMaking2022}, or apply an intervention only for the $k$ top-ranked individuals based on their (predicted) individual outcomes of interest \citep{amarasingheExplainableMachineLearning2023a, kupplerFairPredictionsJust2022b}. The last approach reflects real-world resource constraints typical in the public sector, for example a limited number of staff and financial resources. However, the link between intended goals of a system, allocation principle and prediction targets is often more complicated than this setup suggests. Often additional goals and information needs to be considered before a final decision is made, such as the opinion of a human decision-maker.

\begin{figure*}[t]
\centering
\includegraphics[width=\textwidth]{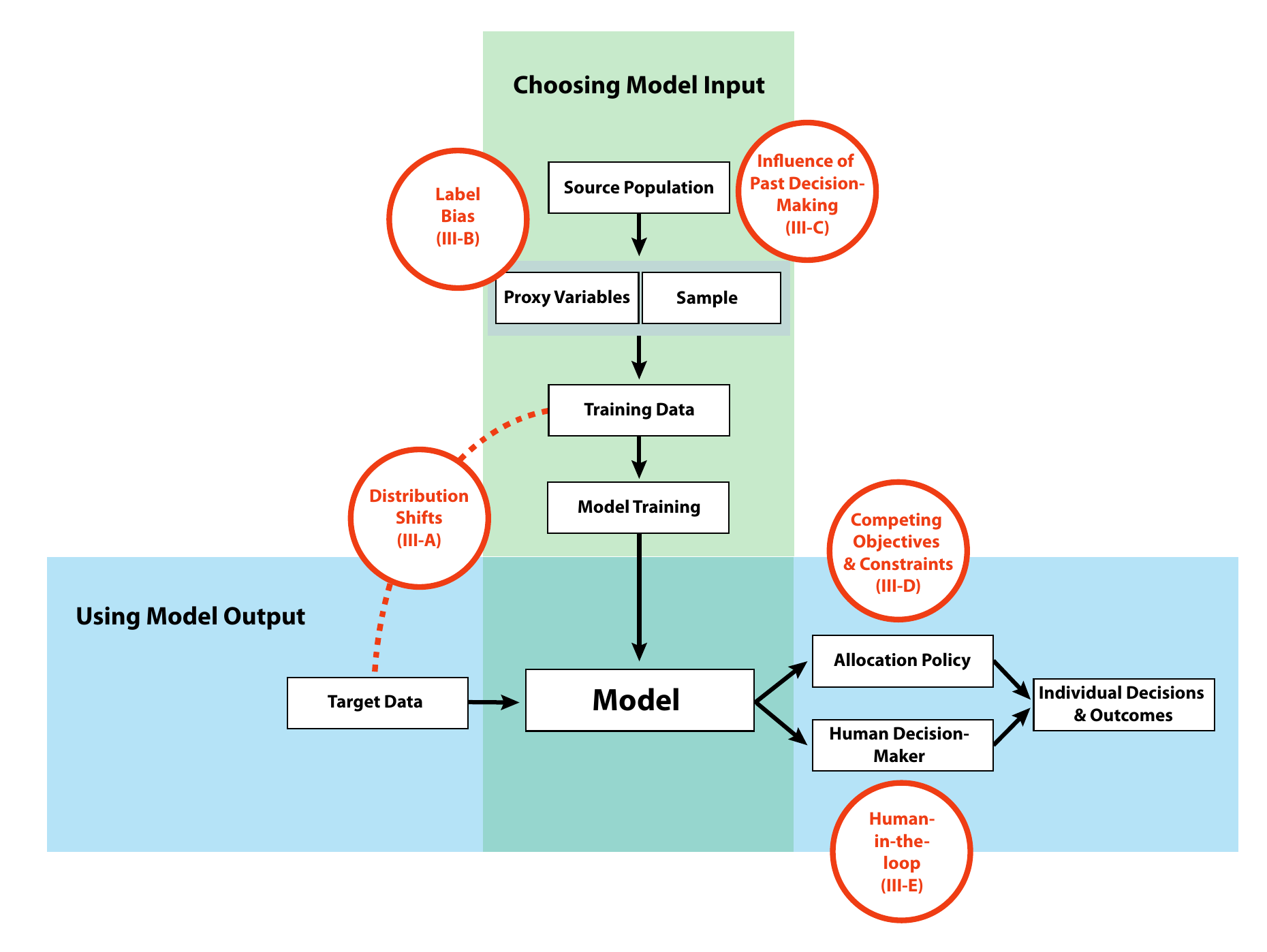}
\caption{Overview of the Primary Technical Challenges at the Intersection of Public Sector Decision-Making and Machine Learning. The challenges (highlighted in red) are positioned along the ML pipeline, with emphasis on data collection and model training (green) and model deployment to support decision-making (blue). For the sake of clarity, some overlapping challenges and connections have been omitted, such as the possible influence of decision outcomes on future data.}
\label{fig: pipeline}
\end{figure*}

In the following subsections, we discuss key challenges associated with both data input and model output that are especially relevant for high-stakes decision-making in the public sector (see Figure \ref{fig: pipeline}). These challenges include considering potential distribution shifts between the model's training and deployment context (Section \ref{sec: distribution shifts}), dealing with proxy variables in complex policy settings (Section \ref{sec: label bias}) and discerning the impact of past decision-making on the data (Section \ref{sec: past decision-making}). We will also discuss the difficulty of handling multiple potentially conflicting goals (Section \ref{sec: competing objectives}), and the role of human decision-makers whose judgment can potentially overrule the recommendations made by an algorithm (Section \ref{sec: human-in-the-loop}).

\subsection{Distribution Shifts}
\label{sec: distribution shifts}
A key challenge when using ML models is to ensure that they perform well in real-world situations. Often, the data used to train and evaluate the model will not fully represent the actual population in the environment where the model will be deployed \citep{kouwIntroductionDomainAdaptation2018, gruberSourcesUncertaintyMachine2023a}. This mismatch between the distribution of training and deployment data is commonly referred to as distribution shift, and can lead to a significant overestimation of a model's performance \citep{kouwIntroductionDomainAdaptation2018, gruberSourcesUncertaintyMachine2023a}. In other words, the model may learn patterns in the training data that do not generalize well to the deployment data, causing it to perform poorly in practice. Distribution shifts are especially challenging in the public sector, where sourcing reliable data can be particularly difficult. Models are often deployed in complex and evolving social contexts, and limited resources, such as a shortage of technical staff \citep{wirtzArtificialIntelligencePublic2019}, make it difficult for public institutions to monitor model performance for unexpected distribution shifts.

There are several types of shifts that can occur \citep{kouwIntroductionDomainAdaptation2018, moreno-torresUnifyingViewDataset2012}. For example, the distribution of input covariates, such as age, income or educational background, might vary if a model is trained in one geographic region but deployed in another. An even harder type of shift to address is when the relationship between input covariates and the outcome changes, making the model's predictions less applicable in the new setting.

Distribution shifts often result from a biased selection of training data \citep{moreno-torresUnifyingViewDataset2012}. For example, it may be more costly to collect relevant data for hard to reach subgroups in the population \citep{tourangeau2019surveying}, leading to them being underrepresented in the data used for training. Selection bias may be introduced through a variety of other mechanisms, for example if past decision policies have led to only certain subgroups receiving an intervention, it will be difficult to assess the interventions' impact for other individuals. 


Even a comprehensive selection of training data does not guarantee long-term robustness. As changes in the deployment environment occur, the initially accurate data may become increasingly outdated, likely causing the performance of a model to degrade over time \citep{moreno-torresUnifyingViewDataset2012}. For instance, labor market characteristics might change over time, making a model trained on older data for predicting unemployment less accurate. When a model is used to inform future decision-making, its continued deployment may itself be a source of distribution shift. For instance, individuals might strategically manipulate attributes that are not causally related to the true outcome but are correlated to improve predictions in their favor, often worsening the model's accuracy in predicting the true outcome of interest \citep{hardtStrategicClassification2016}. This is a known challenge when making use of models to support enforcement decisions in government, such as financial fraud detection. In order to evade detection, certain regulatory subjects will adapt to a given system, thereby requiring continuous updates to maintain effectiveness \citep{engstromGovernmentAlgorithmArtificial2020b}. The performance of a model may also be impacted by more sudden changes in the deployment environment. The introduction of a new policy can influence how new training data is collected and labeled, and unexpected events, such as the COVID-19 pandemic \citep{singhEvaluatingWidelyImplemented2021}, can reduce the prediction accuracy of a model.

One common strategy to keep a ML model up-to-date is to regularly retrain it with new incoming data. However, caution is needed in scenarios where the model's output strongly influences future training data \citep{perdomoPerformativePrediction2020}. Biased initial training data can result in self-fulling prophecies, in which model-informed interventions lead to new biased data that is fed back into the model training. Predictive policing is a canonical example of such a harmful feedback loop, where a higher police presence in neighborhoods classified as high-risk by the model can lead to higher arrest rates (i.e. the proxy variable) independent of the true crime rate \citep{ensignRunawayFeedbackLoops2018}. 

To effectively anticipate distribution shifts, the insight of domain experts and stakeholders will often be key. For example, prior knowledge of which causal relationships between predictors and the outcome of interest are expected to remain invariant \citep{kerriganSurveyDomainKnowledge2021} can facilitate the selection of a dedicated approach to increase robustness under shifts. In general, prediction quality should be continuously monitored to detect signs of worsening model performance. This task goes beyond the technical challenges we will discuss, and necessitates dedicated institutional resources and procedures, such as requiring periodic re-approval of deployed systems \citep{levyAlgorithmsDecisionMakingPublic2021}.

\subsection{Label Bias}
\label{sec: label bias}

Obtaining accurate ground truth data in real-world settings is rarely simple, as the true quantity of interest is often not directly measurable \citep{costonValidityPerspectiveEvaluating2023a,  guerdanGroundLessTruth2023a, barocasFairnessMachineLearning2019}. While challenges in measuring outcomes are not unique to the public sector, they are particularly pronounced in the public sector, where projects often address complex social phenomena that are difficult to quantify such as health, social welfare and education. In contrast, the private sector typically evaluates outcomes using more straightforward metrics like return on investment \citep{wirick2011public}. These difficulties often encourage the use of proxy variables that are more easily available, such as hospitalization records and arrest rates. Similarly, it may take a considerable amount of time before the outcome of interest can be observed; predicting the 10-year risk of cardiovascular events takes at least a decade. Such lag may require the use of short-term outcomes as proxies \citep{atheySurrogateIndexCombining2019a}. 

Using proxy variables can introduce bias into a model. Proxy variables often capture institutional responses rather than the true underlying outcome of interest. For example, using ICU hospitalization as a proxy for COVID-19 severity is an imperfect measure, as ICU admission will depend on other factors like bed availability and other admission criteria. This can be especially problematic if the relationship between proxy and true target varies by protected attributes, such as race and gender \citep{guerdanGroundLessTruth2023a, passiProblemFormulationFairness2019}. \citet{obermeyerDissectingRacialBias2019a} demonstrate that using expected healthcare cost as a proxy for health needs in predictive algorithms can lead to significantly underestimating the risk score of Black patients. This is because Black patients with similar health needs generate fewer medical expenditures compared to white patients. Similar examples can be found in various application contexts, such as judicial bail prediction \citep{fogliatoFairnessEvaluationPresence2020} and lending algorithms \citep{mitchellAlgorithmicFairnessChoices2021}.

Mitigating such biases cannot be achieved by collecting more data; it demands careful consideration of the relationship between the the true label $Y$ and the measured proxy label $\tilde{Y}$ \citep{gruberSourcesUncertaintyMachine2023a, guerdanGroundLessTruth2023a}. Validating the assumptions made about the measurement process may require an evaluation of the proxy variable using external data. For example, in the evaluation of the Allegheny Family Screening Tool, an algorithm designed to aid in child maltreatment hotline screening, researchers utilized data from a pediatric hospital in form of hospitalization records to assess the relationship between the model's risk scores and the occurrence of injury encounters as recorded in the hospital's dataset \citep{vaithianathan2019allegheny, chengHeterogeneityAlgorithmAssistedDecisionMaking2022a}.

\subsection{Past Decision-Making}
\label{sec: past decision-making}
When developing an ADM system, we often encounter scenarios in which the available training data has been influenced by past decision-making \citep{costonCounterfactualRiskAssessments2020b}. For example, a model predicting the risk of job seekers becoming long-term unemployed with the aim of allocating future support programs needs to account for how such programs were distributed in the past. Otherwise the model will likely underestimate the risk of unemployment for individuals that used to receive prioritized support after the decision-making process is altered through the deployment of the model \citep{lenertPrognosticModelsWill2019}. The predictions of such a model would lead to misleading recommendations, since its predictions are valid only under the assumption that the decision-making policies remain unchanged or that the interventions were largely ineffective in the past \citep{dickermanCounterfactualPredictionNot2020}.

In such scenarios, it may be required to explicitly model the effect of interventions by predicting counterfactual outcomes, such as the expected outcome of a medical treatment for a specific individual. However, the estimation of counterfactual outcomes is difficult, as it relies on untestable assumptions due to the limitation of only observing one intervention outcome per individual. Causal modeling requires data on past interventions, specifically which interventions each individual was targeted with, whereas missing treatment data is common in real-world scenarios \citep{kennedyEfficientNonparametricCausal2020, kuzmanovicEstimatingConditionalAverage2023b}. A central challenge in causal modeling are confounding variables, resulting in the group of individuals subjected to a specific intervention exhibiting systematic differences in outcomes compared to the overall population \citep{fernandez-loriaCausalDecisionMaking2022}. Students from a more privileged socioeconomic background may find it easier to enroll in a free tutoring program, but may also tend to perform better on tests due to stronger support networks. This leads to a risk of overestimating the effectiveness of the program for students from the general population.

The canonical way of dealing with such bias are randomized controlled trials (RCTs) \citep{caronEstimatingIndividualTreatment2022}. However, in many high-stakes public sector settings it will be impossible to conduct a RCT due to resource constraints and ethical limitations \citep{caronEstimatingIndividualTreatment2022}. When the efficacy of the interventions is well-established, a randomized study may be hard to justify, as in the case of criminal justice \citep{lakkarajuSelectiveLabelsProblem2017} or child abuse prevention \citep{vaithianathanUsingMachineLearning2021}.

Alternatively, causal outcomes may be estimated from observational data. However, this requires the assumption that relevant confounding variables have been observed, allowing for the disentanglement of past intervention assignment and outcomes. However, some variables may remain elusive \citep{rambachanCounterfactualRiskAssessments2022, lakkarajuSelectiveLabelsProblem2017}, such as the impressions gained by decision-makers from in-person interactions. In situations in which it is difficult to guarantee no unmeasured confounding variables, there still may be ways to estimate the outcome of interest. For instance, \citet{chenLearningSelectiveLabels2023a} and \citet{lakkarajuSelectiveLabelsProblem2017} utilize data from multiple human decision-makers who were randomly assigned to cases to enable estimation.

\subsection{Competing Objectives and Constraints}
\label{sec: competing objectives}
Formalizing the intended policy objectives into a clearly defined allocation principle is difficult, especially when dealing with multiple stakeholder groups, each with their distinct and potentially competing goals and constraints \citep{levyAlgorithmsDecisionMakingPublic2021, mitchellAlgorithmicFairnessChoices2021, passiProblemFormulationFairness2019}. For example, a welfare agency may seek cost-efficient solutions, while ensuring fair decision-making. Similarly, when the IRS decides whom to audit, various objectives come into play, such as maximizing revenue, deterrence, and compliance with institutional and monetary constraints \citep{blackAlgorithmicFairnessVertical2022}. Regardless of the specific context, resource constraints are common in the public sector \citep{amarasingheExplainableMachineLearning2023a}. These constraints may result from limited financial resources or be influenced by institutional factors, such as a limited workforce, legal regulations or external political considerations. 

Predictive systems, however, typically encourage a more limited scope by estimating only one relevant factor \citep{mitchellAlgorithmicFairnessChoices2021}. This singular focus may introduce omitted-payoff bias, a situation where a model target captures only a subset of critical objectives and constraints, potentially reducing the real-world utility of the system \citep{kleinbergHumanDecisionsMachine2018}. For example, the IRS disproportionately audits low-income owners compared to their high-income counterparts, despite higher misreporting of tax liability among the latter group \citep{blackAlgorithmicFairnessVertical2022}. While auditing low-income individuals is more cost-efficient, it can exacerbate social inequalities. This problem of narrow focus becomes especially pronounced when human decision-makers have fewer opportunities to incorporate additional considerations into the decision-making process and rely on the model's predictions too heavily.

Therefore, effort must be made to translate multiple policy goals and constraints into explicitly defined objectives for the ADM system \citep{coyleExplainingMachineLearning2020a, mitchellAlgorithmicFairnessChoices2021}. The exact choice of the prediction target often represents a policy choice because it can have profound downstream impacts that should not solely be the responsibility of ML developers \citep{levyAlgorithmsDecisionMakingPublic2021, passiProblemFormulationFairness2019}. For instance, in the IRS example, shifting the prediction target from the probability of misreporting to predicting misreported income leads to a significantly more equitable distribution of audits, even without explicitly enforcing fairness constraints \citep{blackAlgorithmicFairnessVertical2022}. Integrating multiple goals into a system typically requires making explicit tradeoffs between different objectives and constraints. A familiar example of competing objectives during model development is that accuracy has to be sacrificed to enforce fairness constraints \citep{kozodoiFairnessCreditScoring2022, blackAlgorithmicFairnessVertical2022} or enhance model interpretability \citep{murdochDefinitionsMethodsApplications2019}. In public policy, one common approach to assess competing goals is performing a cost-benefit analyses, valuing different impacts and objectives in monetary terms \citep{boardmanCostBenefitAnalysisConcepts2018}. A similar approach in model design might permit the combination of multiple objectives into a single loss function. However, assigning a monetary value to different potentially incommensurable impacts or goals is not always straightforward, resulting in critiques of this utilitarian approach to decision-making \citep{hwangCostbenefitAnalysisIts2016}.

Clearly specifying the optimization targets of an ADM system is a delicate process that carries the risk of distorting the originally intended goals \citep{levyAlgorithmsDecisionMakingPublic2021}. This risk is especially pronounced when some policy goals are easier to formalize than others, prompting the oversimplification of complex issues through an algorithmic lens \citep{levyAlgorithmsDecisionMakingPublic2021}. Stakeholders' preference for cost-effective, straightforward solutions and easily measurable prediction targets may exacerbate this problem \citep{barocasFairnessMachineLearning2019}. Nevertheless, decisions must be made, and the growing use of ML in the public sector will likely require new dialogues among stakeholders, while also providing an opportunity to make the weighting and tradeoffs between policy objectives more explicit and transparent than in the past \citep{coyleExplainingMachineLearning2020a, levyAlgorithmsDecisionMakingPublic2021}.

\subsection{Human-in-the-Loop}
\label{sec: human-in-the-loop}

Automated systems alone often cannot meet all the criteria necessary for real-world deployment, such as ensuring reliability under unexpected conditions, transparency and accountability. This makes integrating human decision-makers with algorithmic systems a central concern in the public sector, where systems inform high-stakes decisions and need to comply with complex regulatory frameworks. \cite{mitrou2022} highlight the need for human discretion and oversight when systems continuously learn on (biased) historical data, face competing objectives and values, or need to meet accountability obligations. These concerns are often reflected in legal frameworks like the EU's proposal for AI regulation, which stresses the importance of human oversight for high-risk systems in Article 14 \citep{ProposalREGULATIONEUROPEAN2021}. Thus, the goal is not to replace humans with ADM systems but to assist public decision-makers in their tasks \citep{enarssonApproachingHumanLoop2022}.  

In such scenarios, humans must maintain the final say, making it important to consider how they interpret model outputs and integrate them into their decision-making process. This shifts the focus from simply building the most accurate prediction models to evaluating the consequences of providing specific model recommendations to human decision-makers \citep{fernandez-loriaRejoinderCausalDecision2022, vodrahalliUncalibratedModelsCan2022a}. This introduces new challenges for model developers, who need to ensure that the output of an ADM system can be effectively used by a human decision-maker.

Studies have shown that users are often hesitant to follow recommendations of a predictive model, a phenomenon known as algorithm aversion \citep{de-arteagaCaseHumansintheLoopDecisions2020a, dietvorstOvercomingAlgorithmAversion2018, dietvorstAlgorithmAversionPeople2015a}. This lies in contrast with the opposing tendency of automation bias, when humans excessively rely on a machine's suggestion \citep{goddardAutomationBiasSystematic2012, de-arteagaCaseHumansintheLoopDecisions2020a}. Ongoing research into how humans interact with algorithmic decision-making systems \citep{chugunovaPuttingHumanLoop2023} highlights how these challenges differ based on application contexts and user characteristics. For instance, \citet{chengHeterogeneityAlgorithmAssistedDecisionMaking2022a} demonstrate how less experienced child welfare hotline call workers tend to rely more on an algorithmic risk score than senior workers. Such insights need to guide model development to ensure that the technical design aligns with user requirements and preferences, enabling human decision-makers to make optimal choices. This can involve complex tradeoffs; for example, \citet{chugunovaPuttingHumanLoop2023} illustrate that allowing users to modify algorithmic recommendations increases their willingness to adopt them, but tends to decrease decision accuracy. 

For the interaction between human decision-makers and ML models to work, model predictions and its functioning must be comprehensible for human users \citep{yeomansMakingSenseRecommendations2019, nouraniEffectsMeaningfulMeaningless2019, amarasingheExplainableMachineLearning2023a}. For instance, \citet{lebovitzEngageNotEngage2022} show how opaque ML models make it more difficult for medical professional to effectively use them for diagnosis. Many methods have been proposed to make ML models interpretable and explainable, with comprehensive overviews available in \citet{bellePrinciplesPracticeExplainable2021a, molnar2022, doshi-velezRigorousScienceInterpretable2017a, murdochDefinitionsMethodsApplications2019, rudinInterpretableMachineLearning2022}. One of the reasons why these approaches vary widely is because they have very different conceptions of what constitutes an understandable explanation of the output of a model. \citet{amarasingheExplainableMachineLearning2023a} establish an initial taxonomy linking public policy use cases with existing explainable ML approaches. Moving forward, they stress the need to rigorously evaluate explainable ML methods in real-world problem contexts to ensure their effectiveness in achieving real policy goals and in aiding domain experts. Research from fields such as psychology, cognitive sciences and philosophy may help in the task of creating explanations that are helpful to human users \citep{millerExplanationArtificialIntelligence2019b}. This requires careful investigation of various challenges, such as identifying situations where model explanations may be harmful due to information overload \citep{poursabzi-sangdehManipulatingMeasuringModel2021}, or when users may take advantage of increased transparency to exploit a system \citep{molnar2022}. Similarly, misleading explanations can be used to manipulate users and unjustifiably increase trust in a system \citep{lakkarajuHowFoolYou2020a}.

Providing uncertainty estimates for individual predictions can be critical for enhancing human decision-making based on algorithmic recommendations \citep{bhattUncertaintyFormTransparency2021b}. Uncertainty estimates allow human decision-makers to assess the reliability of a prediction, and when it is necessary to manually intervene \citep{gruberSourcesUncertaintyMachine2023a, shalitCommentaryCausalDecision2022}. This is particularly relevant due to the human tendency to rapidly lose trust in algorithmic systems upon observing errors, despite the algorithm's superior overall performance compared to human decision-makers \citep{dietvorstAlgorithmAversionPeople2015a}. Transparently communicating uncertainty to model users can therefore be a key element for building trust, which also illustrates the need for research into effective communication of probabilities to humans \citep{bhattUncertaintyFormTransparency2021b, vodrahalliUncalibratedModelsCan2022a}.

\section{Expanding the ADM Toolkit: Choosing the Target Estimand}
\label{sec: frameworks}

In Section \ref{sec: challenges}, we outlined several challenges that could threaten the intended functioning of an ADM system using supervised ML models to inform public sector decision-making. These challenges highlight several limitations of solely relying on a traditional supervised ML framework in model design \citep{wangPredictiveOptimizationLegitimacy2023}. In response to these limitations, there have been calls to move beyond purely predictive modeling towards ML methodology that more directly centers around the goal of decision-making \citep{hullermeierPrescriptiveMachineLearning2021a}. This involves a shift in perspective from solely focusing on achieving accurate predictions to a more holistic modus operandi centered on selecting a modeling approach that can best inform the decision-making for a given policy goal and application context.

\begin{figure*}[t]
\centering
\includegraphics[width=\textwidth]{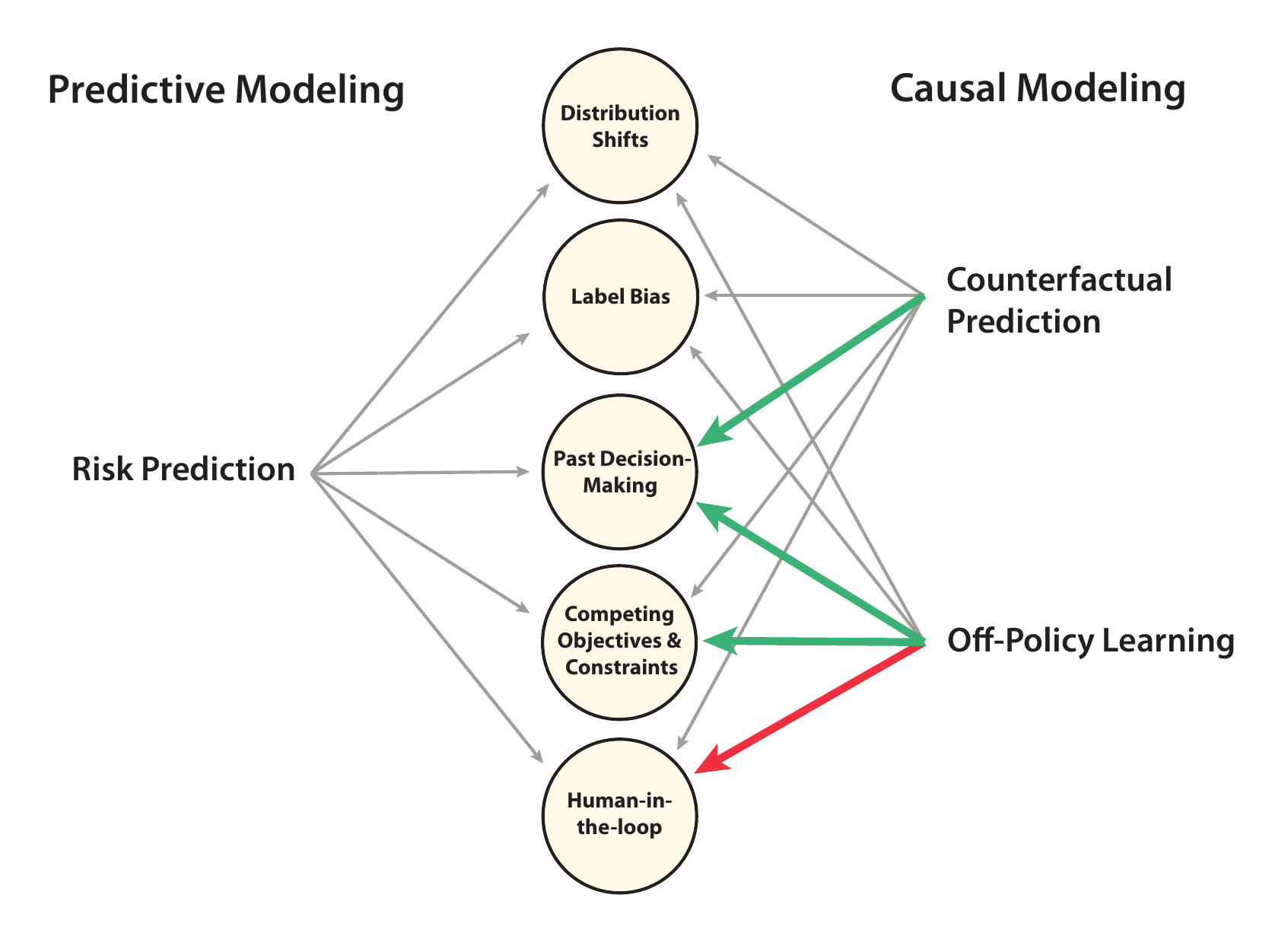}
\caption{Comparing ML Frameworks for Public Sector Decision-Making. Green lines indicate where a causal ML framework is potentially better at addressing a challenge, red lines highlight additional difficulties, and gray lines represent the baseline difficulty of using standard predictive modeling.}
\label{fig: transition}
\end{figure*}

To illustrate this shift, we will discuss three distinct modeling frameworks, starting with standard risk prediction, which is commonly used in ADM systems, and then move to explore two additional causal modeling frameworks. Standard (risk) prediction (Section \ref{sec: risk-prediction}) focuses on estimating outcomes based on historical data without explicitly considering causality. Counterfactual modeling (Section \ref{sec: counterfactual-prediction}) extends this approach by estimating causal outcomes of different hypothetical decisions, directly addressing issues such as the influence of past decision-making on the available outcome data. Lastly, policy learning (Section \ref{sec: policy-learning}) aims to directly learn decision policies that maximize a predefined overarching utility, offering a practical approach to optimize a decision policy within the constraints of real-world scenarios (Section \ref{sec: competing objectives}). Figure \ref{fig: transition} visually compares how well counterfactual prediction and policy learning address the challenges we have discussed in the context of standard prediction.

First, our goal is to examine the implicit and explicit assumptions underlying each modeling framework. This involves addressing two questions: 1) whether the target estimand in each approach is sufficiently linked to the decision-making process, meaning it would genuinely aid in making informed decisions. Understanding these connections is complex, as the guiding principles of an ADM system can be ambiguous, even when specific goals are in place. For example, public employment agencies often seek to allocate resources to job seekers at higher risk of long-term unemployment. However, this objective might stem from either a belief in the effectiveness of early interventions for high-risk job seekers or the notion that high-risk job seekers inherently deserve more support \citep{desiereUsingArtificialIntelligence2021}. Such ambiguity can present difficulties, complicating the choice of the appropriate target estimand, as a need-based distribution necessitates different modeling considerations than an approach focused on the most efficient allocation of interventions. 2) whether estimation is feasible, and what external assumptions are necessary to ensure the accuracy of such estimates. We will discuss these questions for each framework, allowing decision-makers to assess the validity of each approach for their application context.

We will then discuss how the (remaining) challenges outlined in Section \ref{sec: challenges} can be addressed within each framework. We will present methodological advancements specific to each modeling approach that can help overcome the discussed challenges.  While some challenges may be common across all frameworks, others might be more pronounced in specific modeling approaches. Specifically, we focus on distribution shifts to ensure robustness across deployment environments, uncertainty quantification as a key building block for generating trustworthy predictions for humans, and multi-objective optimization to manage tradeoffs between competing objectives.

While we highlight three central modeling approaches, it is important to note that we do not address modeling approaches for every potential decision-making setting. Scenarios involving continuous interventions, interventions across time, sequential data, or interventions that simultaneously target multiple outcomes require other specialized approaches, which are beyond the scope of this paper \citep{hullermeierPrescriptiveMachineLearning2021a,linScopingReviewCausal2021, acharkiComparisonMetaLearnersEstimating2023, vangelovenPredictionMeetsCausal2020}.

\subsection{Risk Prediction}  
\label{sec: risk-prediction} 

In practice, ADM systems commonly involve optimizing predictive models for the estimation of individual outcomes used as decision-criteria \citep{wangPredictiveOptimizationLegitimacy2023}. While predictive ML models are relatively straightforward to set up and train compared to causal models, relying on predictions as proxies for causal outcomes in decision-making runs the risk of generating misleading recommendations \citep{atheyPredictionUsingBig2017, costonCounterfactualRiskAssessments2020b, vangelovenPredictionMeetsCausal2020, wangPredictiveOptimizationLegitimacy2023}. Nevertheless, in certain scenarios, risk predictions may still serve as a useful proxy for decision-making \citep{kleinbergPredictionPolicyProblems2015, guerdanGroundLessTruth2023a, fernandez-loriaCausalDecisionMaking2022}. This is the case if the prediction target is still helpful with regards to the chosen allocation principle. For example, if
the objective is to intervene only in the top-$k$ of individuals, and the ranking induced by the non-causal predictions aligns with that of the causal outcomes, an accurate estimate of the intervention effects may not be critical \citep{fernandez-loriaCausalDecisionMaking2022}.

The validity of such an approach hinges on external assumptions about the relationship between prediction proxies and causal effects. For example, if the predicted outcome is unaffected by interventions, but remains correlated with the causal effects, it can provide valuable information for the allocation strategy, even if it does not correspond directly to the target quantity being optimized. For instance, \citet{kleinbergPredictionPolicyProblems2015} discuss how predicting the mortality risk of patients during the next 1-12 months can be a helpful proxy variable for deciding which patients should not undergo hip and knee replacement surgeries. They argue that patients at risk of death during the months after surgery would not live long enough for the benefits of the surgery to outweigh its costs. Additionally, they assume that the surgery will not significantly impact the mortality risk after the first month, making it possible to determine the optimal intervention — whether to exclude a patient from the surgery or not — based on the predicted risk alone \citep{kleinbergPredictionPolicyProblems2015}.

However, settings in which a practical proxy for the intervention outcome exists may not be common in practice \citep{wangPredictiveOptimizationLegitimacy2023}, because the connection between predicted risk scores and causal effects is rarely straightforward. For example, the predicted risk of death alone would not be sufficient to decide which patients should be first considered for surgery, because the benefits and potential complications of the treatment will probably vary among patients with the same risk score \citep{atheyPredictionUsingBig2017, wangPredictiveOptimizationLegitimacy2023}. Similarly, the Austrian Public Employment Services used a predictive model to assess the risk of long-term unemployment among job seekers with the goal of allocating interventions on this basis \citep{allhutterAlgorithmicProfilingJob2020}. This model divided job seekers into three risk groups. Medium risk individuals were prioritized for support, while high and low-risk individuals were given limited access to labor market programs. However, relying on risk scores to determine an efficient intervention assignment is questionable, as the effectiveness of labor market programs often varies among individuals \citep{cockxPriorityUnemployedImmigrants2023}, even those with the same risk score.

Risk prediction is centered in standard supervised ML methodology, aiming to estimate the statistical relationship between individual covariates $X$ and outcomes $Y$ by learning a prediction function $f: \mathcal{X} \rightarrow \mathcal{Y}$ from a set of observed training data $\mathcal{D} = \{(X_i, Y_i)\}^n_{i=1}$. Even assuming that these predictions provide useful information for the decision-making process, several general threats to the validity of using such a model, as discussed in the previous sections, remain. In the following sections, we will explore methods relevant for describing and tackling these challenges within the realm of risk prediction. This discussion will also lay the groundwork for addressing these challenges in the contexts of counterfactual prediction and policy learning.

\subsubsection{Distribution Shifts, Selection Bias and Label Bias}
Most supervised learning models assume that training and deployment data follow the same distribution. However, in many real-world scenarios we may encounter a distribution shift between the training and deployment environment, necessitating the development of reliable models capable of handling and mitigating such differences \citep{duchiLearningModelsUniform2021}. Training models that remain valid under distribution shift often requires assumptions about the expected type of shift \citep{davidImpossibilityTheoremsDomain2010}. As outlined in Section \ref{sec: distribution shifts}, we will predominantly focus on covariate, label and concept shifts \citep{moreno-torresUnifyingViewDataset2012, quinonero-candelaDatasetShiftMachine2008}. Additionally, we will explore label bias as a type of distribution shift introduced through the use of proxy variables, and consider shifts in time caused by non-stationary environments. We outline central research streams below -- for more in-depth introductions to transfer learning, domain adaptation and out-of-distribution generalization approaches, see \citet{kouwIntroductionDomainAdaptation2018} and \citet{zhouDomainGeneralizationSurvey2022}.

The survey research literature is an invaluable resource for understanding and systematizing different error sources in the data collection process that may surface as distribution shifts downstream. With their inherent focus on valid population inference, concepts such as under-coverage (relevant subpopulations cannot be reached with a data collection schema) and non-response (potentially selective non-participation of relevant subpopulations) extend beyond the traditional survey setting and can help to systematize deficits in the training data in relation to the target population. Error frameworks such as the Total Survey Error \citep{Groves2010, Biemer2010} and its extensions \citep{Sen2021} have been proposed to systematically trace errors along the data collection and processing pipeline that can accumulate in misrepresentation issues. Related work proposes strategies for improving inference from data that do not adequately represent the target population of interest \citep{Cornesse2020, yang2020statistical}. In this context, pseudo-weighting approaches \citep{Elliott-Valliant2017, Valliant-Dever2011} are employed to match the potentially biased source data to some known reference distribution, which resembles methodology from the domain adaptation literature (see below) and similarly connects to concepts in causal inference \citep{Mercer2017}. As a recent example of cross-disciplinary work in this context, \citet{kimUniversalAdaptabilityTargetindependent2022} draw on the multicalibration framework from algorithmic fairness \citep{hebert-johnsonMulticalibrationCalibrationComputationallyIdentifiable2018} to learn prediction functions that are universally adaptable to unknown deployment shifts.

Domain adaptation techniques in the ML literature aim to construct a model that performs well in a setting different from but related to the one it was trained on \citep{kouwIntroductionDomainAdaptation2018, hedegaardSupervisedDomainAdaptation2021}. Unsupervised domain adaptation methods only make use of unlabeled target data to adjust the training data so it better aligns with the deployment distribution \citep{shimodairaImprovingPredictiveInference2000, subbaswamyUnifyingCausalFramework2022}. For example, when a clinical risk prediction tool is deployed in a new hospital, a complete dataset may not be available to re-train the model for the new location. However, it might still be possible to adjust for potential covariate or label shift using unlabeled patient data, assuming that the underlying mechanisms between covariates and outcomes remain invariant. For example, the relationship between diseases and symptoms would not be expected to change between hospitals \citep{liptonDetectingCorrectingLabel2018a}.

Given such data many domain adaptation methods involve importance-weighting, which makes use of the density ratio $w(X) = p_{\mathcal{T}}(X)/p_{\mathcal{S}}(X)$ or class proportions $w(Y) = p_{\mathcal{T}}(Y)/p_{\mathcal{S}}(Y)$ to adjust the loss function \citep{shimodairaImprovingPredictiveInference2000, kouwIntroductionDomainAdaptation2018}. Estimating these weights makes it possible to express the target risk relative to the source distribution $R_\mathcal{T}(\mathcal{L}) = \mathbb{E}_{\mathcal{T}}[\mathcal{L}(X, Y)] = \mathbb{E}_{\mathcal{S}}[w(X, Y)\mathcal{L}(X, Y)]$ which then can be minimized \citep{fangRethinkingImportanceWeighting2020}. Various strategies can be used to estimate the importance weights, such as logistic regression \citep{bickelDiscriminativeLearningCovariate2009a}, kernel density estimation \citep{yuAnalysisKernelMean2012a}, kernel mean matching \citep{quinonero-candelaCovariateShiftKernel2009} and KL-divergence minimization \citep{sugiyamaDirectImportanceEstimation2007}. 

However, weighting methods struggle in settings with limited and complex source data, frequently resulting in high variance estimates, and depend on data being available from the target domain of interest \citep{fangRethinkingImportanceWeighting2020, kouwIntroductionDomainAdaptation2018, liuRobustClassificationSample2014}. Distributionally robust methods offer an alternative approach by providing worst-case guarantees. They often involve minimax estimation, which seek to minimize the loss under the least favorable distribution shift \citep{kouwIntroductionDomainAdaptation2018, subbaswamyUnifyingCausalFramework2022, duchiLearningModelsUniform2021, wenRobustLearningUncertain2014}. 

In addition to the distribution shifts discussed so far, label bias and label noise can present significant challenges, especially given the prevalent use of proxy labels for decision-making. This bias arises when a model is trained not on the true latent label $Y$ of interest but on an erroneous proxy label $\tilde{Y}$. The label bias quantifies the difference between the true distribution of interest $p_T(Y|X)$, and the distribution $p_S(\tilde{Y}|X)$ estimated from the proxy labels \citep{gruberSourcesUncertaintyMachine2023a}. This shift can be characterized by a label corruption process or measurement error model, which describes the probability of a true label $Y$ being recorded as a proxy label $\tilde{Y}$ \citep{gruberSourcesUncertaintyMachine2023a, fangRethinkingImportanceWeighting2020, daiLabelBiasLabel2020}.

Various approaches have been devised to mitigate label noise and measurement error. For instance, \citet{natarajanLearningNoisyLabels2013} propose an unbiased risk minimization strategy for handling class-conditional $p(\tilde{Y} | Y)$ noise. While such a simplified model of a proxy may be applicable in some settings, practitioners will likely encounter more complex scenarios \citep{chen2021beyond}, such as the measurement error depending on sensitive covariates \citep{wang2021fair, obermeyerDissectingRacialBias2019a}. In some situations, there may be the option to access multiple proxies of the true target of interest. For example, \citet{boeschotenAchievingFairInference2021a} utilize a structural equation model to characterize the relationship between multiple proxies and the unobserved outcome to ensure fair predictions. There has also been research into how label bias interacts with other distribution shifts. For example, \citet{daiLabelBiasLabel2020} propose a joint framework for addressing label bias and label shift, while \citet{yuLabelNoiseRobustDomain2020} investigate the interaction between class-conditional noise and generalized target shifts.

Distribution shifts tend to occur gradually over time \citep{webb2016characterizing}, often triggered by the deployment of the model itself. Addressing such feedback loops and ongoing distribution shifts poses a significant challenge, likely requiring future research into the temporal dynamics of ML-informed decision-making \citep{pagan2023classification}. For example, \citet{perdomoPerformativePrediction2020} introduce a modeling framework that incorporates the potential impact of predictions on the predicted outcome of interest. These predictions are referred to as \textit{performative}, effectively leading to distribution shifts by altering the target distribution in the deployment environment over time. They develop the notion of performative optimality, ensuring that a decision rule minimizes the expected loss with regard to the future target distribution it induces. The specific choice of the loss function can align with different objectives. For instance, one may opt to optimize for a target distribution with mostly favorable outcomes instead of solely focusing on accurate predictions \citep{kimMakingDecisionsOutcome2022}. 

\subsubsection{Uncertainty Quantification}

Accurate uncertainty estimates are key for enabling reliable decision-making systems. For example, they make it possible to determine when a model should refrain from making a recommendation and instead fully defer to a human user \citep{gruberSourcesUncertaintyMachine2023a}. While we highlight selected methods below, we refer the reader to \citep{gruberSourcesUncertaintyMachine2023a, bhattUncertaintyFormTransparency2021b, sullivanIntroductionUncertaintyQuantification2015a, hullermeierAleatoricEpistemicUncertainty2021} for comprehensive reviews of the emerging literature on uncertainty estimation in machine learning. 

In recent years, interest has grown in conformal prediction as a distribution-free and model-agnostic approach to uncertainty quantification for ML models. These characteristics make conformal prediction particularly appealing in many practical scenarios, as no specific assumptions on the model are required, enabling easy implementation for any arbitrary ML model. Instead of providing a point prediction, conformal prediction constructs a set of plausible predictions with respect to a chosen significance level \citep{vovkAlgorithmicLearningRandom2022, angelopoulosGentleIntroductionConformal2021, papadopoulosInductiveConfidenceMachines2002a}. A larger conformal set indicates higher uncertainty in the model's predictions. Conformal prediction requires splitting the data into a training set and an additional holdout dataset, known as the calibration set. 
Alternatively, full conformal prediction does not necessitate dividing the data but is usually computationally more demanding \citep{angelopoulosGentleIntroductionConformal2021}. Conformal prediction relies on exchangeable data, which can not be guaranteed in scenarios involving distribution shifts. However, efforts have been made to extend conformal prediction for such situations, such as making use of weighting methods akin to those discussed in the context of unsupervised domain adaptation \citep{tibshiraniConformalPredictionCovariate2019, barberConformalPredictionExchangeability2023a, NEURIPS2021_0d441de7}.  

As mentioned, uncertainty estimates play a significant role in facilitating cooperative interaction between human users and models, particularly for high-stakes decision-making prevalent in the public sector. For example, \citet{straitouriDesigningDecisionSupport2023a} propose a decision-support framework that makes use of conformal prediction to improve the cooperation between experts and the ML model. Their modeling framework restricts domain experts to choose their prediction from a set of plausible predictions generated by the model, resulting in better performance than relying on the model or the human expert alone.

\subsubsection{Multi-Objective Optimization}
A central challenge for ADM systems is balancing multiple objectives within a constrained outcome space. This often requires making tradeoffs between competing objectives, such as determining the appropriate balance between an equitable distribution of resources and maximizing cost-efficiency. Consequently, they require stakeholder input, further complicating the task by necessitating systems that are sufficiently accessible and interpretable for stakeholders to both make and evaluate these tradeoffs effectively \citep{papalexopoulosEthicsbydesignEfficientFair2022}.

In the context of risk prediction, predictive modeling and decision-making are separated into two distinct steps \citep{elmachtoubSmartPredictThen2022, kupplerFairPredictionsJust2022b}. Initially, a prediction is generated that subsequently gets used to inform a downstream allocation problem. In current ADM systems practice, multi-objective optimization is rarely employed. Typically, problems are cast as single-objective constrained optimization tasks, like finding the best allocation within budget constraints. 
When more complex constraints, different decision criteria and multiple predictions come into play, a conventional approach for formalizing the decision step involves the creation of a scalar utility function that linearly combines different objectives into a weighted sum \citep{keeneyDecisionsMultipleObjectives1993a}. For instance, stakeholders might construct a unified risk score out of multiple criteria that is subsequently employed to prioritize the allocation of resources. However, constructing a joint utility function can be tricky \citep{dasCloserLookDrawbacks1997}, as stakeholders often struggle to determine how to exactly weigh different objectives \citep{hayesPracticalGuideMultiobjective2022, roijersSurveyMultiObjectiveSequential2013, boutilierComputationalDecisionSupport2013}. This difficulty is especially pronounced in risk prediction, where the link between predictions and the expected utility is often not entirely specified. A predicted risk score may only allow for a prioritization of individuals while the exact size of the individual utilities remains unknown. For example, in scenarios where intervention costs vary significantly by individual it might be important to compare the exact magnitude of the guiding utility for each individual intervention, making it problematic if only a ranked list is available. 

A common alternative to defining a utility function a priori is to seek allocations that reside along the Pareto front, which constitutes the set solutions where improving one objective necessarily entails the worsening of another \citep{hayesPracticalGuideMultiobjective2022, debMultiobjectiveOptimisationUsing2011}. For example, \citet{hertweckJusticeBasedFrameworkAnalysis2023} propose a framework to visualize tradeoffs between the utility of the decision-maker and the fairness demands of the decision subject. While approaches like this will still leave stakeholders with difficult value choices, they might aid in making tradeoffs more explicit by focusing the selection on a specific set of allocation policies. Similarly, the notion of multi-target multiplicity describes a scenario in which multiple prediction targets that are all considered to be equally valid operationalizations of the outcome of interest are available \citep{watson-daniels2023Multiplicity}. This makes it possible to explore arbitrary combinations of these targets to arrive at an allocation that maximizes group-level fairness.

On the other hand, secondary objectives and constraints such as ensuring models are fair \citep{hortBiasMitigationMachine2022, hardtEqualityOpportunitySupervised2016, zafarFairnessConstraintsMechanisms2017, 10.1145/3097983.3098095} and interpretable \citep{molnar2022} might already come into play in the modeling process. More efforts are being made to naturally integrate such constraints into the ML pipeline. For example, recent work in Multi-Criteria Auto ML \citep{pfisterer2021multiobjective} proposes a framework where users can iteratively specify tradeoffs between different objectives, such as fairness, accuracy and robustness, to explore subregions of the Pareto front. Automatized modeling procedures of this kind might make it easier to interactively elicit stakeholder preferences.

\subsection{Counterfactual Prediction}
\label{sec: counterfactual-prediction}

The primary goal of any ADM system is to guide decision-making by recommending a particular course of action. Making such recommendations effectively will often involve counterfactual modeling. While non-causal risk predictions can be used as proxies for relevant counterfactual outcomes, they risk being significantly biased, potentially making a more principled approach involving explicit causal modeling preferable. However, as outlined in Section \ref{sec: past decision-making}, a common threat to the validity of causal models is confounding, requiring external assumptions and historical data on intervention assignment to address. This challenge requires careful case-by-case analysis by the model developer and may limit the possibility to make use of counterfactual estimates for decision-making in certain application contexts.

The potential outcomes framework \citep{rubinEstimatingCausalEffects1974} is a prominent approach for framing causal questions. In a binary intervention scenario $\mathcal{T} = \{0, 1\}$, it denotes two potential outcomes $(Y_i(0), Y_i(1))$ for an individual $i$. These outcomes represent the two possible observable outcomes: no intervention ($T_i = 0$) and an intervention ($T_i = 1$). The individual treatment effect $\tau_i$ is then defined as the difference between the potential outcomes $\tau_i = Y_i(1) - Y_i(0)$. 
Estimating potential outcomes and treatment effects from observational data $\mathcal{D} = \{(X_i, T_i, Y_i)\}_{i=1}^{n}$ is challenging, as it is usually only possible to observe one outcome $Y_i = (1-T_i) Y_i(0) + T_i Y_i(1)$ for each individual \citep{kunzelMetalearnersEstimatingHeterogeneous2019}. Consequently, it is common to estimate the expected potential outcomes $\mu_t(x) = \mathbb{E} [Y(t) | X = x ]$ and conditional average treatment effect (CATE) $\tau(x) = \mathbb{E} [Y(1) - Y(0) | X = x]$ for a given covariate vector $X = x$ \citep{vegetabileDistinctionConditionalAverage2021a, kunzelMetalearnersEstimatingHeterogeneous2019}.  We refer to Appendix \ref{appendix: cate estimation} for an overview of relevant ML-based CATE estimation methods.

To link the CATE with a statistical estimand, a set of untestable assumptions is required \citep{kunzelMetalearnersEstimatingHeterogeneous2019, caronEstimatingIndividualTreatment2022, johanssonGeneralizationBoundsRepresentation2022a}. Unconfoundedness $(Y(0), Y(1)) \indep T | X$, requires that potential outcomes are conditionally independent of treatment assignment. Positivity guarantees nonzero propensity scores $0 < P(T = 1 | X = x) < 1$  for all confounders $x \in \mathcal{X}$, meaning that treatment assignment is not fully deterministic. Finally, Stable Unit Treatment Value Assumption (SUTVA) assumes that the outcome of one individual is not affected by the interventions others received, and that there are no different versions of a specific treatment. Under these assumptions it becomes in principle possible to infer $\mu_t(x)$ and $\tau(x)$ from observational data \citep{caronEstimatingIndividualTreatment2022}. Ensuring these assumptions can be difficult, with unmeasured confounding posing a significant risk when aiming for valid counterfactual predictions. 

While the individual treatment effect seems like a natural choice for determining the optimal allocation, there exist various scenarios in which estimating only one expected potential outcome $\mu_t(x)$ may be sufficient to inform the decision-making process \citep{dickermanCounterfactualPredictionNot2020}. These outcomes could, for example, represent the likelihood of abuse if a hotline call is not followed up \citep{costonCounterfactualRiskAssessments2020b} or the risk of death of a patient if no heart transplant is performed \citep{vangelovenPredictionMeetsCausal2020, dickermanCounterfactualPredictionNot2020}. Models that provide such risk assessments align well with allocation principles informed by need-based criteria by identifying individuals at high risk of adverse outcomes. For example, when screening phone calls for potential child maltreatment \citep{vaithianathan2019allegheny}, there exists a moral and legal obligation to investigate high risk cases, regardless of the investigation's likelihood of success \citep{costonCounterfactualRiskAssessments2020b, chouldechovaCaseStudyAlgorithmassisted2018a}. Additionally, in scenarios where one potential outcome is trivially known, only one outcome needs to be estimated. For instance, in judicial bail prediction, individuals for whom bail was denied cannot re-offend before trial \citep{lakkarajuSelectiveLabelsProblem2017}.

While assumptions for causal identification are still necessary to correctly estimate expected potential outcomes, in many scenarios this task may be more feasible than full treatment effect estimation. For example, this might be the case when implementing an intervention that has not been previously deployed, or when data on specific outcomes is generally limited \citep{fernandez-loriaCausalDecisionMaking2022}. Following the discussion on proxies in risk prediction, explicitly modeling the treatment effect might also not be necessary if the relationship between a potential outcome and treatment effect is well-established \citep{fernandez-loriaCausalDecisionMaking2022}. For example, prior knowledge and experiments may indicate that a particular treatment strategy is the most beneficial approach for individuals in a specific risk group \citep{athey2023machine}, allowing us to correctly prioritize individuals based on the estimated baseline risk $Y(0)$ alone \citep{fernándezloría2023causal}.

Compared to CATE estimation, this only requires the estimation of a single outcome regression $\mu_t(x) = \mathbb{E}[Y | X = x, T = t]$. While this simplifies some of the necessary considerations for CATE estimation, caution may still be warranted in low-data settings due to differences in the covariate distribution of the treatment group and overall population, potentially necessitating dedicated approaches to correct this imbalance during estimation \citep{johanssonGeneralizationBoundsRepresentation2022a}. A growing body of recent research focuses on auditing and evaluating counterfactual prediction models of this nature for algorithmic decision-making. For example, \citet{costonCounterfactualRiskAssessments2020b} discuss evaluation fairness metrics and evaluation methods for counterfactual risk modeling. In the domain of clinical risk prediction, a substantial body of literature explores methods for predicting outcomes under specific medical treatments \citep{linScopingReviewCausal2021, vangelovenPredictionMeetsCausal2020, schulamReliableDecisionSupport2017, prosperiCausalInferenceCounterfactual2020}.

In the following, we will discuss approaches to address distribution shifts, uncertainty quantification and multi-objective optimization for CATE estimation. While many of the earlier considerations in the context of risk prediction remain applicable, there are aspects unique to this setting, requiring methods dedicated to tackling the challenges for causal modeling.

\subsubsection{Distribution Shifts, Selection Bias and Label Bias}

The challenge of handling distribution shifts is strongly related to causal estimation, as illustrated by \citet{johanssonLearningRepresentationsCounterfactual2016a}. For example, predicting counterfactual outcomes under no unmeasured confounding corresponds to unsupervised domain adaptation under covariate shift \citep{johanssonGeneralizationBoundsRepresentation2022a}. This is because past decision-making policies often lead to a difference in covariate distribution between the treatment groups and the distribution of the overall population. Several approaches for dealing with shifts when performing CATE estimation have been proposed \citep{johanssonLearningRepresentationsCounterfactual2016a, shalitEstimatingIndividualTreatment2017, assaad21a}. For example, \citet{kuzmanovicEstimatingConditionalAverage2023b} study the problem of inferring CATE in settings in which treatment information is missing for some individuals, a challenge they frame as a covariate shift problem. 

In many practical settings, approaches geared towards guaranteeing robustness to unknown distribution shifts \citep{jeongRobustCausalInference2020} may be particularly relevant, as it can be difficult to anticipate the target population and relevant subpopulations in all possible deployment environments. To tackle this challenge, \citet{kernRobustConditionalAverage2023} introduce an approach for learning robust CATE estimates under unknown external covariate shifts. They achieve this by employing a boosting-style post-processing routine to generate a multi-accurate predictor \citep{kimMultiaccuracyBlackBoxPostProcessing2019}, enabling unbiased predictions in a new deployment setting. 

In a recent study, \citet{guerdanGroundLessTruth2023a} examine the interaction of label bias and counterfactual prediction. They propose a causal framework that describes potential biases introduced by proxy labels, and survey strategies for evaluating the chosen measurement model. There are not many approaches that explicitly deal with measurement error in the context of employing counterfactual models. \citet{guerdanCounterfactualPredictionOutcome2023} develop a framework that accounts for treatment-conditional errors based on the previously discussed approach for correcting class-conditional noise \citep{natarajanLearningNoisyLabels2013}. 

\subsubsection{Uncertainty Quantification}

Recently, conformal prediction has been extended to address individual treatment effect estimation, with a central challenge being that exchangeability of the data can not be guaranteed due the covariate shift between treatment groups and the overall population \citep{alaaConformalMetalearnersPredictive2023a}. \citet{leiConformalInferenceCounterfactuals2021b} propose a solution that makes use of weighted conformal prediction \citep{tibshiraniConformalPredictionCovariate2019} to correct for this shift. They construct prediction intervals for potential outcomes, which are then used to derive intervals for the individual treatment effects. In contrast, conformal meta-learners, as introduced by \citep{alaaConformalMetalearnersPredictive2023a}, offer a framework for directly constructing prediction intervals for pseudo-outcomes of two-stage meta-learners, allowing for conformal prediction for a different class of CATE estimation methods. As in the case of risk prediction, providing a conformal set has the potential to facilitate human and model interaction by guiding a decision-maker towards a set of likely solutions, while still leaving the critical final decision to the human.

\subsubsection{Multi-Objective Optimization}

In principle, the challenge of handling multiple objectives and constraints remains similar when employing risk prediction and counterfactual prediction. In both scenarios, multi-objective optimization typically becomes relevant when determining the downstream allocation after a prediction is generated. However, counterfactual estimates are usually easier to link with the intended guiding utility than non-causal predictions, making it more straightforward to quantify tradeoffs with other objectives. 

Efforts have been made to explicitly integrate CATE estimation and prescriptive optimization within an unified framework, typically with a focus on budget-constrained optimization problems \citep{10.1145/3442381.3450075, 10.1145/3485447.3512103}. Formulating such optimization problems is generally made easier when the expected net benefit can be easily defined, such as maximizing net revenue when allocating tax audits within a fixed budget \citep{blackAlgorithmicFairnessVertical2022}. Similarly, \citet{mcfowland2021prescriptive} present a prescriptive analytics framework that combines randomized experiments, CATE estimation and a subsequent constrained optimization problem to identify the profit-maximizing allocation policy. Crucially, the expected cost of an intervention may not necessarily be known, potentially requiring a separate estimation process. Unlike in the case of generic prediction, only a few studies have attempted to integrate constraints directly into the counterfactual estimation process. Notable examples include \cite{kim2023fair} for CATE estimation and \cite{mishlerFairnessRiskAssessment2021} for counterfactual risk prediction under fairness constraints.

\subsection{Policy Learning}
\label{sec: policy-learning}

The ADM approaches discussed so far entail a two-step process: initially estimating individual outcomes, such as the CATE, and subsequently using these estimates to determine an optimal downstream allocation considering external constraints. However, this means that the target of estimation is only indirectly linked to the underlying policy objective, as an improved prediction may not necessarily enhance the utility of the resulting allocation policy \citep{perdomo2023relative}. While perfect predictions could in theory lead to optimal decision-making, in practice it may sometimes more practical to estimate the allocation policy directly \citep{elmachtoubSmartPredictThen2022, fernandez-loriaCausalDecisionMaking2022}. A wide range of methods have been proposed to optimize the aggregated utility, emphasizing that the primary goal of deploying a statistical targeting system is not accurate prediction alone. 

More specifically, the target of estimation becomes the allocation policy $\pi: \mathcal{X} \rightarrow \{0, 1\}$, directly mapping individual covariates $X_i$ to an intervention. Learning optimal assignment rules has been studied across different disciplines, such as statistical decision theory, economics and operations research \citep{manski2004, kitagawaWhoShouldBe2018, elmachtoubSmartPredictThen2022}. This paper specifically highlights recent literature focusing on learning optimal policies from past observational data using ML methods \citep{atheyPolicyLearningObservational2021, kallusBalancedPolicyEvaluation2018,  hatamyar2023policy, luedtke2016statistical}. Here, we are concerned with off-policy learning, given that on-policy learning may not be suitable for high-stakes settings in the public sector where active experimentation with different decision policies is not possible.

Off-Policy learning usually involves optimizing over a class of policies $\pi \in \Pi$ by defining the aggregated utility of a proposed policy $V(\pi) = \mathbb{E} [Y(\pi(x))]$ as the overall expected outcome if the policy were deployed \citep{atheyPolicyLearningObservational2021}. Finding the optimal policy corresponds to identifying the policy that maximizes the utility, i.e. $\hat{\pi} = \underset{\pi \in \Pi}{\argmax}  \ \hat{V}(\pi)$. When estimating the policy value from observational data, we rely on the same assumptions as those used in CATE estimation to ensure identification, such as unconfoundedness. We refer to Appendix \ref{appendix: off-policy learning} for an overview of off-policy learning methods.

Adopting a population-level perspective and directly optimizing for the best allocation policy can come with several benefits. First, such an approach may aid in the natural integration of downstream constraints, for example by limiting the class of policies $\Pi$ under consideration to these that can feasibly be implemented. For instance, policy learning enables the exclusion of decision policies that make use of specific covariates \citep{kallusMoreEfficientPolicy2021, atheyPolicyLearningObservational2021}, such as sensitive attributes like race and gender, or features susceptible to individual manipulation potentially leading to distribution shifts after deployment. While these variables may be required to address confounding during estimation, we can ensure that the decision-making does not rely on them by constraining the class of allowed policies. Second, estimating and optimizing the policy value $V(\pi)$ for a constrained set of policies is a distinct and potentially easier estimation task compared to predicting individual-level treatment effects \citep{kallusMoreEfficientPolicy2021, lechnerCausalMachineLearning2023a}. In general, precise estimation of individual-level outcomes may not always be a prerequisite for determining the optimal policy, as an erroneous prediction may not necessarily lead to an erroneous decision \citep{fernandez-loriaCausalDecisionMaking2022}. 

The feasibility of employing policy learning also depends on whether the goal of the modeling process is a fully automated decision system or providing recommendations to human decision-makers. As discussed in Section \ref{sec: human-in-the-loop}, fully formalizing the connection between model output and decision-making can be challenging due to the involvement of human decision-makers who may want to integrate external information and can overrule the model's recommendation \citep{shalitCommentaryCausalDecision2022}. This complication adds nuance to the discussion about choosing the most appropriate target estimand and modeling framework. For example, a human decision-maker might find a CATE estimate more trustworthy and more suitable for individual decision-making, as opposed to a fully defined allocation policy \citep{costonCounterfactualRiskAssessments2020b}. Conversely, a well-defined policy class could also be restricted to policies that can be easily interpreted by users, such as decision trees \citep{atheyPolicyLearningObservational2021}.

In the next section, we will highlight relevant work extending policy learning to tackle distribution shifts, uncertainty quantification and multi-objective optimization. While policy learning shares many similarities with CATE estimation, it still involves a distinct target estimand and estimation strategies, requiring tailored approaches to this setting. Research at the intersections of policy learning and the aforementioned challenges is still in early stages, but there have been some promising developments in the recent past. 

\subsubsection{Distribution Shift, Selection Bias and Label Bias}

As described, a significant challenge in managing distribution shifts for ADM systems lies in precisely specifying the anticipated changes from the historical environment to the future deployment environment. For example, the data-generating mechanism may change over time, but the exact nature of this shift is usually hard to predict. Addressing this challenge, \citet{siDistributionallyRobustPolicy2020, siDistributionallyRobustBatch2023} propose an algorithm for distributionally robust policy learning under unknown covariate and concept shift. Their approach involves maximizing the worst-case policy value over all environments within a specific distance to the training environment. 
By choosing their preferred distance, decision-makers can manage their risk aversion before deploying a policy \citep{siDistributionallyRobustBatch2023}. Building on this work, \citet{kallusDoublyRobustDistributionally2022a} incorporate doubly-robust methods, removing the need to assume that the historical assignment policy is known, which is often unavailable when relying on observational data. 
Instead of ensuring robustness under arbitrary distribution shifts, it may also be helpful to focus on specific types of shifts, potentially simplifying the integration of domain knowledge. For example, \citet{hattGeneralizingOffpolicyLearning2022} develop a framework for learning worst-case policies that generalize under distributional shifts resulting from an unknown selection bias.

\subsubsection{Uncertainty Quantification}

After selecting a policy for deployment, especially in high-stakes settings, reliable uncertainty estimates become important to guarantee the policy's reliability. Uncertainty quantification in off-policy learning often involves estimating bounds for the expected aggregate utility of the policy under hypothetical deployment \citep{NEURIPS2022_cc84bfab}, for example as seen in \citet{wang2017optimalpolicyeval}. However, in many scenarios there may arise the need to quantify the uncertainty of outcomes at the individual level. For instance, a policy that appears to lead to a positive aggregate utility may still be deemed unacceptable if the variability in outcomes for certain subgroups is overly large. Recent works in off-policy evaluation have investigated the application of conformal prediction to construct prediction intervals. Similar to CATE estimation, a critical challenge for conformal off-policy prediction lies in guaranteeing exchangeability of the data. \citet{pmlr-v206-zhang23c} and \citet{NEURIPS2022_cc84bfab} have proposed approaches that make use of weighted conformal prediction to address the shift between training data and deployment environment, allowing for the reliable estimation of prediction sets.

\subsubsection{Multi-Objective Optimization}

To generalize the policy value for multiple objectives, one can consider the weighted sum of utilities resulting from various individual outcomes and external objectives. For example, in scenarios with individually varying intervention costs, it may be possible to define a net-monetary benefit, that is subsequently used as the target outcome in the policy value \citep{xu2022estimating}. Alternatively, a decision-maker may want to enforce an overarching constraint, such a limited budget, as part of the policy optimization problem \citep{10.1111/biom.13232, sun2021empirical}. However, if the relevant constraint needs to be adjusted regularly, such approaches may lead to significant computational costs \citep{sun2024treatment}. \citet{sun2024treatment} propose learning an individual-level priority score that directly encodes the cost-benefit ratio, which can subsequently easily be used to rank individuals for intervention under varying resource constraints.  
Furthermore, off-policy learning methods that optimize within a constrained class of policies \citep{atheyPolicyLearningObservational2021, kallusMoreEfficientPolicy2021} have the advantage that secondary constraints can also be encoded through restricting the class of allowed policies. For example, \citet{frauenFairOffPolicyLearning2023a} propose a policy learning that restricts the policy class to those respecting fairness constraints.

In practice, decision-makers frequently encounter scenarios with multiple objectives that are not easily expressed as constraints or monetary monetary costs. As described, identifying the set of Pareto-optimal models may allow stakeholders to effectively explore tradeoffs between objectives. \citet{rehill2022policy} propose a multi-objective Bayesian optimization approach for off-policy learning, utilizing proxy models to efficiently construct the Pareto-Frontier, enabling a human user to better evaluate the consequences of different weightings between objectives.

\section{Towards a Decision-Centric ML Toolkit in the Public Sector}
\label{sec: toolkit}

Making productive use of ML for public sector decision-making is a complicated task, requiring careful alignment of policy objectives and model development. First, we set out to explore challenges faced when deploying ML for public sector decision-making. Specifically, we focused on challenges arising from a misalignment between policy objectives and technical design, such as when assumptions about the training data do not match the intended application context. We identified and discussed five key challenges in Section \ref{sec: challenges}, highlighting potential limitations of solely relying on the standard supervised ML paradigm (see Figure \ref{fig: transition}). In response to these limitations, we examined alternative modeling frameworks, specifically counterfactual prediction and off-policy learning. Each framework comes with its own set of distinct advantages and is potentially better suited to overcome some of the challenges. To choose between these frameworks, a model developer should consider two key questions. 1) How will the estimated quantity be helpful in decision-making? 2) What is necessary to ensure unbiased estimation?

We observe that targets that are easier to estimate, often require more assumptions about their utility in the decision-making process. Even with perfect knowledge of these targets, they might need to be combined with domain knowledge to be informative for the decision-maker. Conversely, estimating causal outcomes requires more assumptions for causal identification, but might offer more actionable insight for decision-makers. We provide a distilled version of our presentation of different modeling approaches and their links to policy objectives and decision-making in Table \ref{tab: summary}, with the aim of providing guidance for discussions on which modeling theme is most suitable in a given scenario.

\begin{table}[t]
  \centering
  \includegraphics[width=\textwidth]{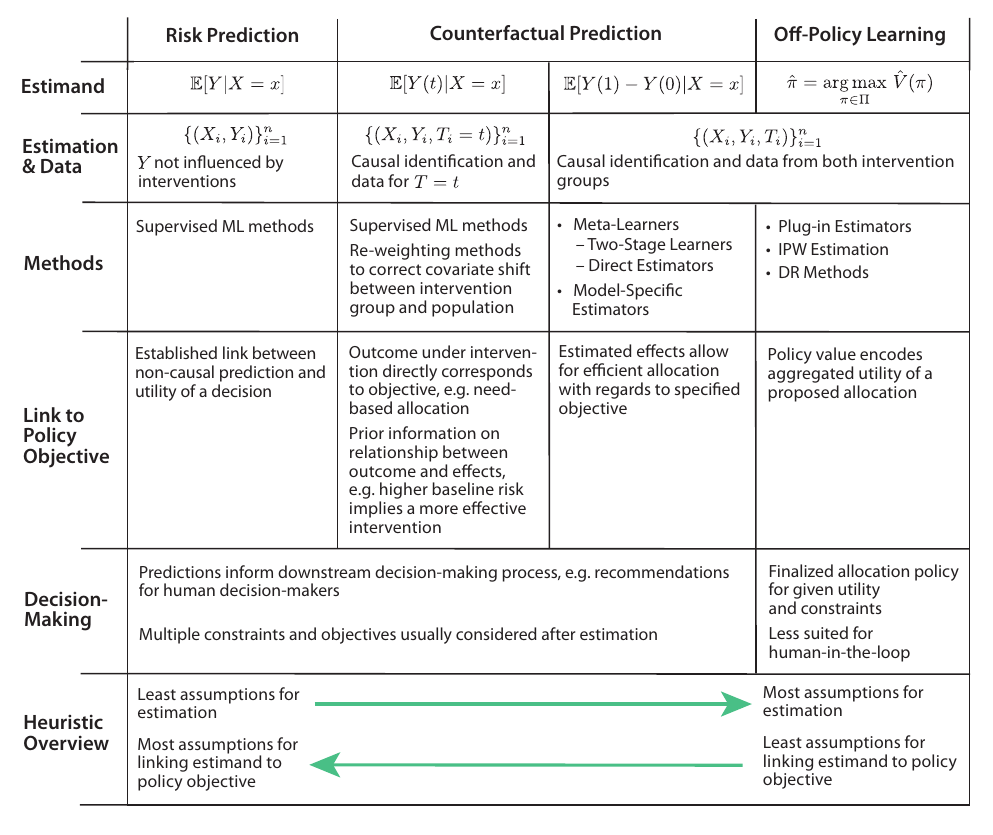}
  \caption{Overview of different (causal) ML frameworks for Algorithmic Decision-Making. The validity of each approach is highly context-dependent, and requires careful evaluation of the available data, decision-making processes and policy objectives.}
  \label{tab: summary}
\end{table}

For example, traditional (risk) prediction methods, while requiring fewer assumption during the estimation process, are not well-suited to estimate counterfactual outcomes, which are often the true target of interest for decision-makers. This limitation often requires additional assumptions about how the predictions relate to the decision-maker's objective. Consider the previously discussed medical scenario as an example for such an assumption: the decision-maker assumes that a higher mortality risk in the coming years might make a knee replacement less beneficial, while also assuming that the mortality risk is not significantly impacted by the surgery itself. Counterfactual prediction can potentially support a variety of decision-makers goals with fewer explicit assumptions about how the goals of the decision-maker align with the target of estimation. For example, consider a public employment service aiming to target support measures to job seekers with a high risk of becoming long-term unemployed. Similarly, reliable estimates of the heterogeneous causal effects of a support program would aid a decision-maker in matching individual's to the most effective program. However, counterfactual estimation is more involved than standard (risk) prediction, relying on external assumptions to ensure causal identification. Policy learning is explicitly integrated into the decision-making process by optimizing a predefined utility function to directly estimate an allocation policy. Such an end-to-end approach allows for a more straightforward integration of constraints and additional objectives. However, it might struggle in scenarios in which human decision-makers play a crucial role. In such settings, decision-makers might prefer individual-level estimates as recommendations to guide their own judgments. We present three examples inspired by real-world use cases in Figures \ref{fig:risk_prediction}, \ref{fig:counterfactual_prediction} and \ref{fig:policy_learning}, each focusing on risk prediction, counterfactual prediction and policy learning respectively. They summarize some of the key questions practitioners need to address to ensure that the selected approach fits the intended application context.

Certain challenges such as the influence of past decision-making are inherently addressed by causal modeling frameworks. To tackle the remaining challenges, we have compiled for each modeling approach a selection of methods to address them, as detailed in the previous section and summarized in Table \ref{tab:challenges and methods} in the appendix. Our goal was to identify methods that are applicable across various ML models within each respective modeling framework, to keep most of our discussion model agnostic and relevant across many application contexts. Unsurprisingly, there is generally less research addressing specific challenges within newer causal modeling frameworks. This gap presents a compelling direction for future research to explore which mitigation strategies from standard supervised ML could be extended to the causal setting.

In recent years, several studies have applied causal ML frameworks to practical applications in the public sector. For example, these modeling approaches have been used and evaluated for the optimal allocation of development aid \citep{kuzmanovic2024causal}, allocation of medical preventive care \citep{doi:10.1287/msom.2021.0251}, child welfare hotline screening \citep{costonCounterfactualRiskAssessments2020b}, and assignment of training programs to job seekers \citep{cockxPriorityUnemployedImmigrants2023}.

\section{Discussion}
\label{sec: discussion}

We analyzed challenges that result from a misalignment between the (technical) assumptions made during model design and the intended policy goals. Each challenge can lead to harmful unintended consequences, impacting the individuals affected by the decisions and potentially undermining the legitimacy of the system.

\begin{itemize}
    \item \textit{Distribution Shifts} occur when the data used to train the ML model does not reflect real-world conditions, causing the performance of the model to decline. Such shifts can lead to misclassifications of individuals and create harmful feedback loops that reinforce erroneous predictions.
    \item \textit{Label Bias} can happen when a ML model relies on proxy variables to estimate hard to measure outcomes. If these proxies are biased and primarily reflect institutional practices instead of of the true target, they can systematically disadvantage certain groups, leading to unfair predictions and decision outcomes.
    \item \textit{Past Decision-Making} often influences the available training data. If we do not explicitly account for these past interventions, the predictions of the ADM system may become outdated once new decision-making practices are implemented. For example, a model might underestimate the risk for individuals who previously received support, potentially leading to harm if future allocations fail to consider this.
    \item \textit{Competing Objectives and Constraints} can complicate the formalization of policy goals in an ADM system, as predictive systems often focus on singular, clearly defined objectives. This narrow focus can introduce omitted-payoff bias, such as when optimizing for cost-efficiency disproportionately harms marginalized groups.
    \item \textit{Human-in-the-Loop} is an important component of ADM systems, because automated systems alone often do not meet all necessary requirements for real-world deployment. However, interactions between models and human decision-maker can introduce complications and biases. For example, an accurate prediction may not be helpful if case workers lack trust due to unclear communication of model uncertainties.
\end{itemize}

In this work, we found that standard machine learning approaches are not necessarily well-suited for the public sector context, requiring model designers to expand their toolkit to effectively address these issues. We discussed different technical solution strategies in detail and provided guidance on choosing between alternative modeling frameworks - including predictive and causal modeling - to tackle these challenges. However, to do so effectively, the technical model design needs to be guided in close collaboration with domain experts. Each challenge we presented is embedded in complex, changing social contexts, where purely technical solutions often fall short. The right technical design choice often has no clear or definite answer and can not be left to the model developer alone. For example, the implicit assumptions a model developer makes about the data distribution, the causal structure of the problem, or how different objectives should be represented in the system can greatly affect the validity of the resulting system. However, these assumptions often need to be informed domain-specific knowledge, necessitating insights from policy makers, social scientists and other stakeholders involved. In Table \ref{tab:external input}, we summarize technical solution strategies for each challenge and highlight the points where external input may be central to ensure that technical design choices remain aligned with real-world policy objectives. However, engaging and collaborating with stakeholders is often difficult in practice. Participatory design of AI systems is often mentioned as an important approach, but can be difficult to implement effectively. By specifying the points along the ML pipeline where stakeholders collaboration is crucial for supporting technical design decisions, we hope to guide these efforts and make participatory approaches more actionable \citep{delgado2023}.

\begin{table}[!htbp]
\footnotesize
\centering
\begin{tabular}{p{3.5cm}|p{6cm}|p{6cm}}
\centering\textbf{Challenge} & \centering\textbf{Technical Solution Strategies} & \multicolumn{1}{c}{\textbf{Stakeholder Input}} \\
\hline
\textit{Distribution Shifts} & \compress
\begin{itemize}[leftmargin=*]
        \item Apply domain adaptation methods using data from the deployment environment
        \item Use distributionally robust optimization for worst-case guarantees
        \item Implement continuous monitoring of input data and model performance
    \end{itemize} &  \compress
\begin{itemize}[leftmargin=*]
        \item Collaborate with domain experts and data providers to anticipate changes in the deployment environment (e.g. changing regulations and user behavior)
        \item Engage stakeholders to determine acceptable risk tolerance
        \item Involve decision-makers to determine relevant evaluation metrics for ongoing monitoring
        \item Identify vulnerable and hard-to-reach sub-populations
    \end{itemize} \\
\hline
\textit{Label Bias} &\compress \begin{itemize}[leftmargin=*]
        \item Construct measurement error models for selected proxy variables
        \item Validate chosen proxy variables using external data sources and additional variables
\end{itemize} & \compress \begin{itemize}[leftmargin=*]
        \item Collaborate with domain experts to understand how proxy variables map to the true concepts of interest
        \item Identify societal biases that may impact the proxy-target relationship
    \end{itemize}  \\
\hline
\textit{Past Decision-Making} & \compress \begin{itemize}[leftmargin=*]
        \item Identify suitable (causal) estimands and estimation strategies, such as CATE estimation, counterfactual prediction or policy learning
\end{itemize}& \compress \begin{itemize}[leftmargin=*]
        \item Validate link that the chosen (causal) estimand is able to inform the decision-making process
        \item Make use of domain expertise to inform assumptions necessary for causal identification, such as gathering knowledge on past decision-making criteria and processes
    \end{itemize} \\
\hline
\textit{Competing Objectives and Constraints} & \compress \begin{itemize}[leftmargin=*]
        \item Integrate external constraints into model design and allocation procedure (e.g. using model multiplicty and constrained optimization)
        \item Identify solutions along the Pareto frontier to enable decision-makers to manage tradeoffs
    \end{itemize} &\compress \begin{itemize}[leftmargin=*]
        \item Elicit preferences from decision-makers to quantify tradeoffs between different objectives and constraints
        \item Collaborate with stakeholders to identify objectives not fully captured by the ADM system
    \end{itemize} \\
\hline
\textit{Human-in-the-Loop} & \compress \begin{itemize}[leftmargin=*]
        \item Use uncertainty quantification (e.g. conformal prediction) and explainable ML methods to provide guarantees to decision-makers and enhance transparency
\end{itemize}    &\compress \begin{itemize}[leftmargin=*]
    \item Understand how model outputs will be interpreted and used by decision-makers, considering user background and workflows
    \item Regularly gather feedback from end-users to improve model integration
\end{itemize}    \\
\hline
\end{tabular}
\caption{Overview of key challenges in ADM systems and technical solution strategies, focusing on the role of domain expertise in guiding technical design choices.}
\label{tab:external input}
\end{table}

However, our focus does not cover all potential issues related to the use of predictive algorithms in the public sector. While ensuring that a system accurately reflects the goals of decision-makers is a prerequisite for ethical and reliable use, this alone is not sufficient. We do not explicitly discuss ethical, legal and broader societal challenges. Even if a system functions perfectly according to its intended goals, it can still lead to adverse outcomes. For example, this can happen if a decision-maker does not prioritize fair treatment of sensitive subgroups as an explicit design goal \citep{barocasFairnessMachineLearning2019}. 

The intention behind this paper was to clarify the assumptions behind different (predictive) modeling approaches, and help practitioners identify where common technical assumptions may not hold true in the public sector context. We see this as a first step towards the development of a robust methodological framework for constructing and maintaining ADM systems in the public sector that includes both best practices for practitioners and an up-to-date array of technical approaches. While we have made some inroads here by selecting and consolidating relevant theoretical advancements, a critical need for more research to connect these methods with real-world policy use cases remains. As illustrated by the methods and challenges presented here, achieving this goal will likely require bringing together researchers from various disciplines to develop systems that genuinely improve decision-making in tangible ways. It will require a shift in perspective away from technical approaches purely centered around predictive optimization and towards ones that explicitly incorporate decision-making and its impact into the modeling framework.

Effectuating this shift will involve opening up the ML pipeline to external input at significant points. On the one hand, this will require the development of effective strategies for engaging stakeholders and harnessing their expertise, particularly in integrating domain knowledge into the model-building process and eliciting values and objectives from decision-makers. At the same time, more work will have to be done to figure out how the development of ADM systems interacts with and can be embedded into institutional processes and structures. The prospect of this shift might seem daunting at first. It will involve establishing both technical and institutional frameworks that enable the development of successful ADM systems. However, this path also holds the potential to transform our public policy processes in a positive way. It could lead to the establishment of new standards of transparency and the explication of previously implicit goals, as well as facilitating the development of new structures to integrate domain knowledge and involve important stakeholders. Thus, bridging the gap between explicit formalization and nuanced policy requirements could not only unlock the potential of successful ML applications in the public sector, but also lead to a public sector that is more understandable, open to scrutiny and thus accountable. 

\section{Conclusion}
\label{sec:conclusion}

In this paper, we analyzed misalignments between the assumptions underlying ML models and the realities of public sector decision-making. We isolated and discussed five central challenges: distribution shifts, label bias, the influence of past decision-making, competing objectives and constraints and the integration of human decision-makers. We demonstrated how misalignment can lead to unreliable and harmful predictions, potentially causing systems to fail in achieving the intended policy goals and undermining the legitimacy of the decision-making process. Through our analysis, we concluded that many assumptions commonly made in the implementation of ML models do not hold in  complex, evolving decision-making environments. In response, we argue for a shift in modeling efforts from focusing solely on predictive accuracy to improving decision-making outcomes. We presented alternative modeling approaches, including causal machine learning methods including counterfactual prediction and policy learning, which may be better suited to inform decision-making. We also provided guidance on selecting the appropriate modeling strategy by clarifying the assumptions underlying these approaches. Model developers should carefully consider how the estimated quantities can guide decision-making and whether unbiased estimation is possible given available data and external assumptions. Additionally, we summarized technical solutions to the discussed challenges, such as distributionally robust optimization, uncertainty quantification and multi-objective optimization within each modeling framework. Finally, we found that selecting the right methods and frameworks requires external input from domain experts and stakeholders to ensure that the implicit assumptions made by model developers align with the specific problem setting.

\section*{Acknowledgment}

This work is supported by the DAAD programme Konrad Zuse Schools of Excellence in Artificial In- telligence, sponsored by the Federal Ministry of Education and Research, by the Baden-Württemberg Foundation under the grant “Fairness in Automated Decision-Making - FairADM” and by the Volkswagen Foundation, grant “Consequences of Artificial Intelligence for Urban Societies (CAIUS)”. We express our gratitude to Nanina Föhr, Felix Henninger, Patrick Schenk, and Daria Szafran for their valuable feedback and support, and to Bernd Fitzenberger, for shaping our thinking about applications of prediction models in the public sector.

\newpage
\bibliography{bibliography}

\newpage
\appendix

\section{Methodological Approaches for ADM Systems in Public Sector Decision-Making}
\label{appendix: table}

\renewcommand{\arraystretch}{1.3}
\begin{table}[H]
  \centering
  \scriptsize
  \resizebox{\textwidth}{!}{%
  \begin{tabular}{p{\dimexpr 0.15\textwidth-2\tabcolsep}|
                  p{\dimexpr 0.3\textwidth-2\tabcolsep}|
                  p{\dimexpr 0.3\textwidth-2\tabcolsep}|
                  p{\dimexpr 0.3\textwidth-2\tabcolsep}}

    &  \footnotesize \textbf{Risk Prediction} & \footnotesize \textbf{Counterfactual Prediction} &   \footnotesize \textbf{Off-Policy Learning} \\
    \hline
    \textbf{Distribution Shifts} & 
    \compress
    \begin{itemize}[leftmargin=*]
        \item Biases in Data Collection \citep{Groves2010, Biemer2010, Cornesse2020, yang2020statistical, Elliott-Valliant2017, kimUniversalAdaptabilityTargetindependent2022}
        \item Domain Adaptation Using Data from Deployment Environment \citep{kouwIntroductionDomainAdaptation2018, shimodairaImprovingPredictiveInference2000, fangRethinkingImportanceWeighting2020}
        \item Worst-Case Guarantees \citep{duchiLearningModelsUniform2021, wenRobustLearningUncertain2014, zhang2020coping}
        \item Shifts induced by Model Predictions \citep{pagan2023classification, perdomoPerformativePrediction2020, kimMakingDecisionsOutcome2022}
    \end{itemize}
    & 
    \compress
    \begin{itemize}[leftmargin=*]
        \item Covariate Shift between Intervention Groups \citep{johanssonGeneralizationBoundsRepresentation2022a, shalitEstimatingIndividualTreatment2017, assaad21a}
        \item Worst-Case Guarantees \citep{kernRobustConditionalAverage2023, jeongRobustCausalInference2020}
    \end{itemize}
    & 
    \compress
    \begin{itemize}[leftmargin=*]
        \item Worst-Case Guarantees \citep{siDistributionallyRobustBatch2023, siDistributionallyRobustPolicy2020, kallusDoublyRobustDistributionally2022a, hattGeneralizingOffpolicyLearning2022}
    \end{itemize}\\
    \hline
    \textbf{Label Bias} & 
    \compress
    \begin{itemize}[leftmargin=*]
        \item Class-Conditional Label Noise \citep{natarajanLearningNoisyLabels2013, yuLabelNoiseRobustDomain2020}
        \item Feature-Conditional Label Noise \citep{wang2021fair, chen2021beyond} 
        \item Multiple Proxy Variables \citep{boeschotenAchievingFairInference2021a}
    \end{itemize}
    &  
    \compress
    \begin{itemize}[leftmargin=*]
        \item Counterfactual Prediction under Measurement Error \citep{guerdanCounterfactualPredictionOutcome2023, guerdanGroundLessTruth2023a}
    \end{itemize}   
    & No dedicated methods specific to policy learning have been identified. Methods from other frameworks may be applicable. \\
    \hline
    \textbf{Past Decision-Making} 
    & Unbiased estimation not possible if the outcome was influenced by interventions.
    & Requires causal identification. See Appendix \ref{appendix: cate estimation} for an overview of ML-based CATE estimation methods.
    & Requires causal identification. See Appendix \ref{appendix: off-policy learning} for an overview of off-policy learning methods.
    \\
    \hline
    \textbf{Competing  Objectives \& Constraints} 
    & 
    \compress
    \begin{itemize}[leftmargin=*]
        \item Scalar Utility Functions \citep{keeneyDecisionsMultipleObjectives1993a, boutilierComputationalDecisionSupport2013}
        \item Solutions along the Pareto Frontier \citep{hertweckJusticeBasedFrameworkAnalysis2023, pfisterer2021multiobjective} and Model Multiplicity \citep{watson-daniels2023Multiplicity}
        \item Specific Model Constraints, such as Fairness \citep{hortBiasMitigationMachine2022, hardtEqualityOpportunitySupervised2016} and Interpretability \citep{molnar2022}
    \end{itemize}   
    & 
    \compress
    \begin{itemize}[leftmargin=*]
        \item Budget-Constrained Allocation \citep{10.1145/3442381.3450075, 10.1145/3485447.3512103, mcfowland2021prescriptive}
        \item Fairness Constraints \citep{kim2023fair, mishlerFairnessRiskAssessment2021}
    \end{itemize}   
    & 
    \compress
    \begin{itemize}[leftmargin=*]
        \item Budget-Constrained Allocation \citep{xu2022estimating, 10.1111/biom.13232, sun2021empirical}
        \item Solutions along the Pareto Frontier \citep{rehill2022policy}
        \item Specific Model Constraints \citep{atheyPolicyLearningObservational2021}, such as Fairness \citep{frauenFairOffPolicyLearning2023a}
    \end{itemize}   
    \\
    \hline
    \textbf{Uncertainty Estimation for Human-in-the-Loop}
     & 
    \compress
    \begin{itemize}[leftmargin=*]
        \item Model agnostic Uncertainty Estimation with Conformal Prediction \citep{papadopoulosInductiveConfidenceMachines2002, angelopoulosGentleIntroductionConformal2021, straitouriImprovingExpertPredictions2023a}
    \end{itemize}   
     & 
    \compress
    \begin{itemize}[leftmargin=*]
        \item Model agnostic Uncertainty Estimation for CATEs with weighted Conformal Prediction \citep{leiConformalInferenceCounterfactuals2021b, tibshiraniConformalPredictionCovariate2019} and conformal meta-learners \citep{alaaConformalMetalearnersPredictive2023a}
    \end{itemize}   
    & 
    \compress
    \begin{itemize}[leftmargin=*]
        \item Uncertainty in Policy Value \citep{wang2017optimalpolicyeval} and Conformal Off-Policy Evaluation for Outcomes \citep{NEURIPS2022_cc84bfab}
    \end{itemize}
    \\
    \bottomrule
  \end{tabular}
  }
  \caption{Overview of Methodological Approaches to Address Key Challenges of ADM Systems in the Public Sector in Risk Prediction, Counterfactual Prediction, and Off-Policy Learning.}
  \label{tab:challenges and methods}
\end{table}

\begin{figure*}[!h]
\centering
\includegraphics[width=\textwidth]{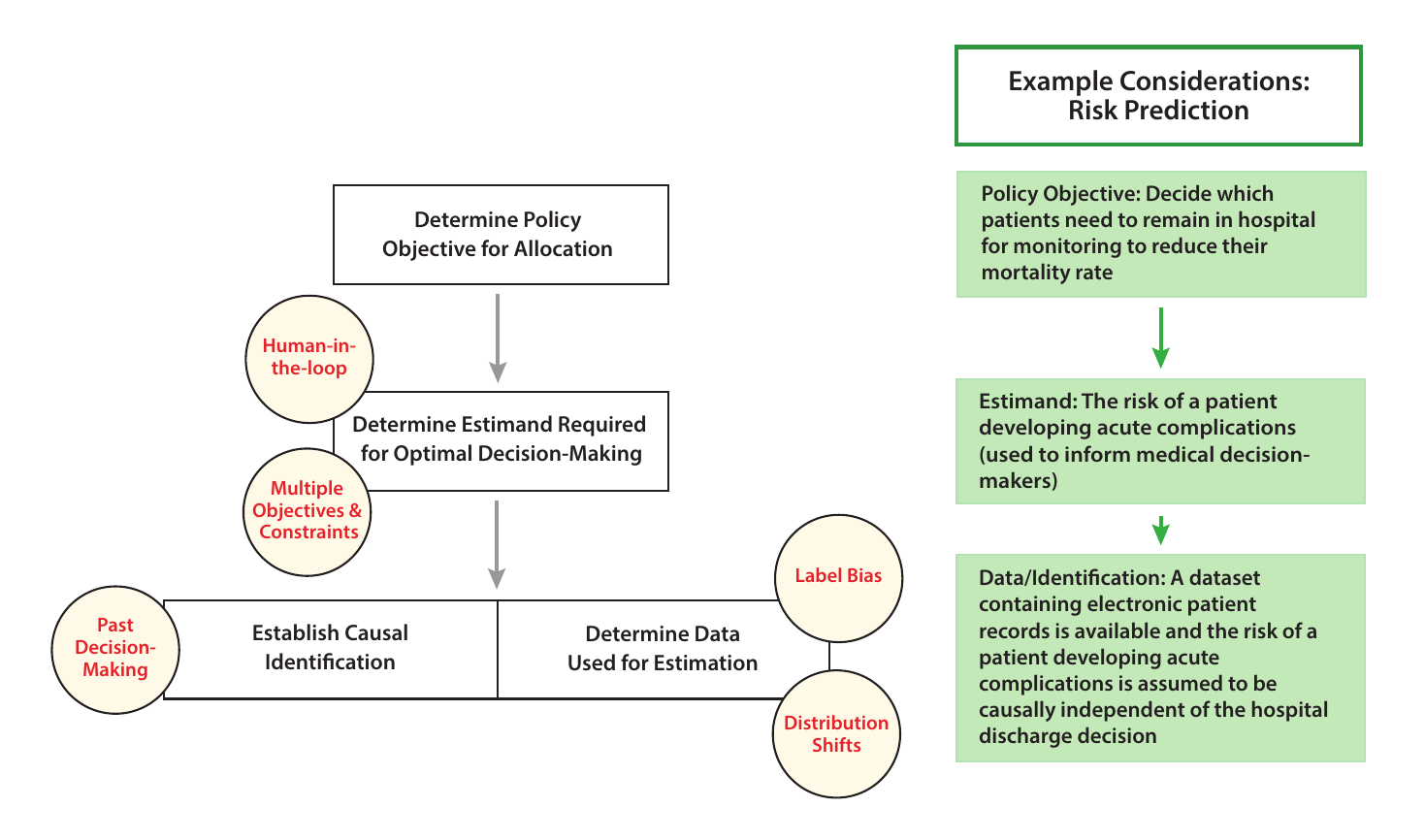}
\caption{Key Questions for Policy Makers in Selecting Risk Prediction as the Modeling Approach. Example inspired by algorithmic predictions of acute gastrointestinal bleeding \citep{NEURIPS2023_fb44a668}}
\label{fig:risk_prediction}
\end{figure*}

\newpage

\begin{figure*}[!h]
\centering
\includegraphics[width=\textwidth]{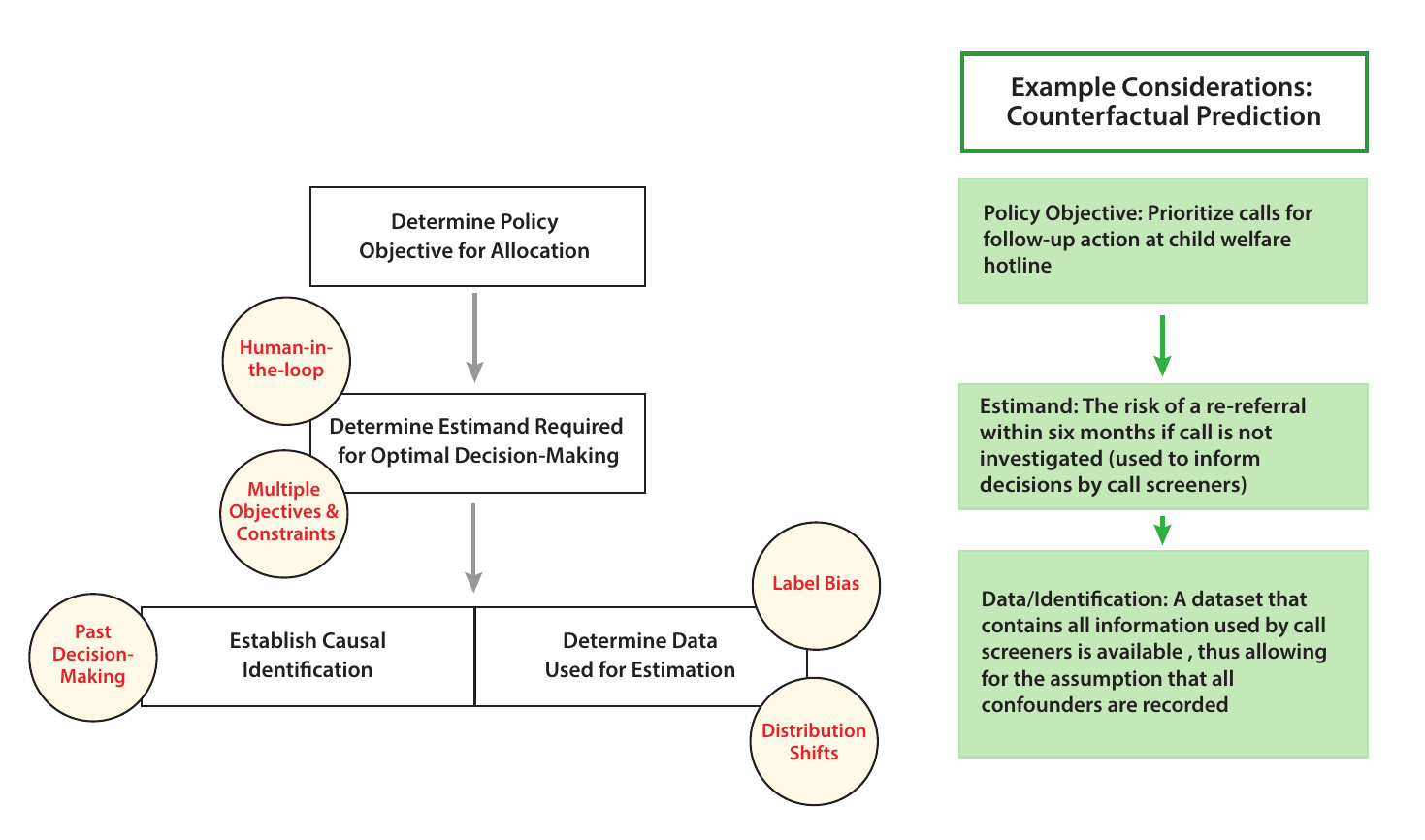}
\caption{Key Questions for Policy Makers in Selecting Counterfactual Prediction as the Modeling Approach. Example inspired by child abuse hotline screening in Allegheny County \citep{costonCounterfactualRiskAssessments2020b, chouldechovaCaseStudyAlgorithmassisted2018a}}
\label{fig:counterfactual_prediction}
\end{figure*}

\begin{figure*}[!h]
\centering
\includegraphics[width=\textwidth]{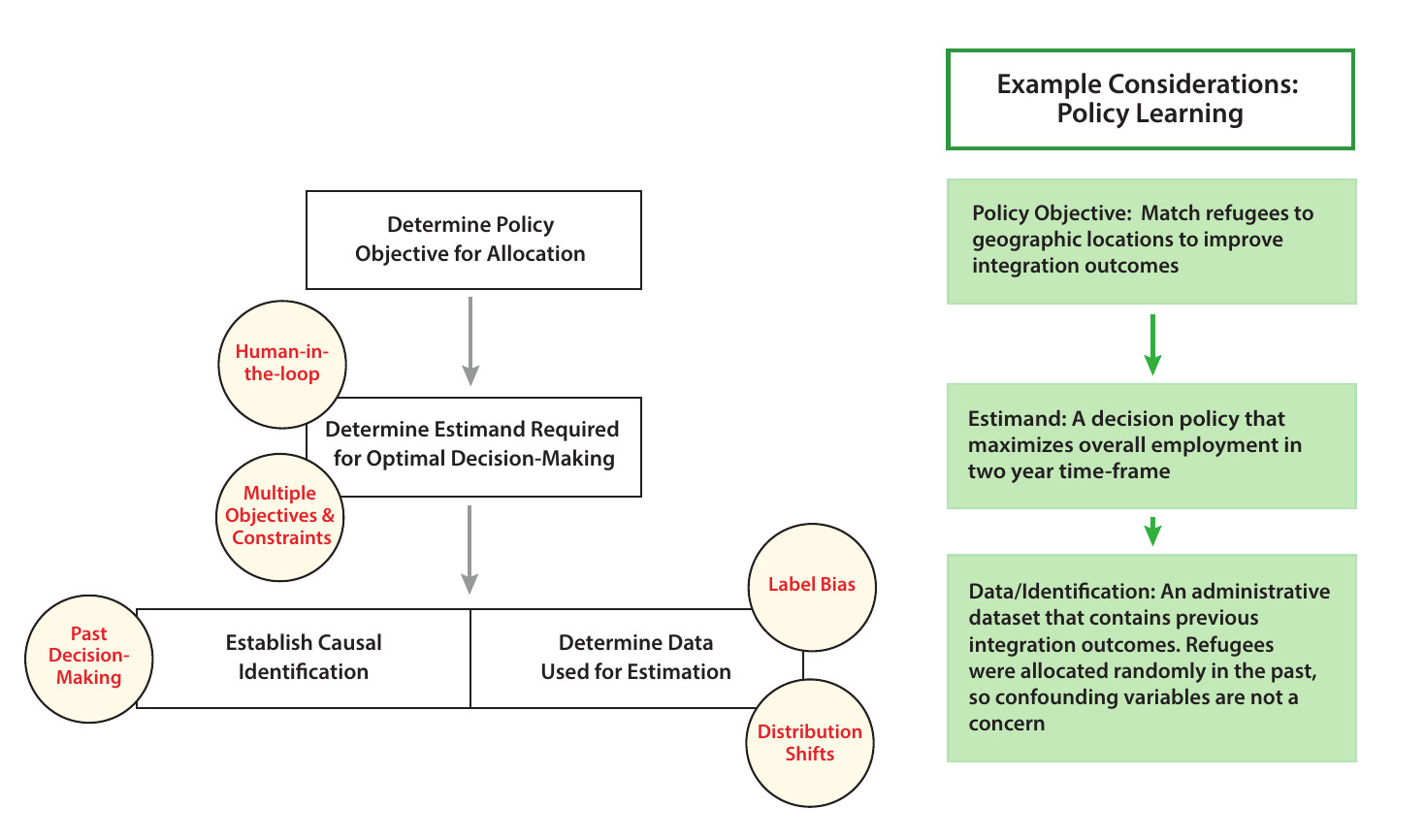}
\caption{Key Questions for Policy Makers in Selecting Policy Learning as the Modeling Approach. Example inspired by geographical matching of refugees to improve integration outcomes \citep{doi:10.1126/science.aao4408}}
\label{fig:policy_learning}
\end{figure*}

\newpage

\section{CATE Estimation Methods}
\label{appendix: cate estimation}
In recent years, several CATE estimation strategies have emerged, many of which employ nonparametric ML regression models to estimate the relationship between covariates, outcome and intervention \citep{lechnerCausalMachineLearning2023a, caronEstimatingIndividualTreatment2022}. ML models are generally well-suited for inferring complex non-linear relationships and handling a larger number of covariates, which can be vital for capturing heterogeneous effects. Model-agnostic meta-algorithms for CATE estimation enable the use of an arbitrary ML model as a base learner, such as random forests and neural networks \citep{kunzelMetalearnersEstimatingHeterogeneous2019, caronEstimatingIndividualTreatment2022, curthNonparametricEstimationHeterogeneous2021}. 

One class of meta-learners initially aims to estimate both expected outcome functions $\mu_t(x)$, and computing the CATE as the difference between the estimates of these functions \citep{curthNonparametricEstimationHeterogeneous2021}. For example, S-learners treat the treatment indicator as an additional feature and estimate the potential outcomes with a single outcome regression $\mu(x, t) = \mathbb{E} [Y | X = x, T = t]$. T-learners use ML models to estimate $\mu_0(x) = \mathbb{E} [Y | X = x, T = 0]$ and $\mu_1(x) = \mathbb{E} [Y | X = x, T = 1]$ separately. However, both approaches can introduce significant bias, particularly when dealing with imbalanced treatment groups \citep{johanssonGeneralizationBoundsRepresentation2022a, nieQuasiOracleEstimationHeterogeneous2020a, caronEstimatingIndividualTreatment2022, kunzelMetalearnersEstimatingHeterogeneous2019}. 

Alternative methods directly estimate the CATE function by first constructing pseudo-outcomes of the treatment effects \citep{kunzelMetalearnersEstimatingHeterogeneous2019, caronEstimatingIndividualTreatment2022, curthNonparametricEstimationHeterogeneous2021}. One prominent variant is the X-learner \citep{kunzelMetalearnersEstimatingHeterogeneous2019}, an extension of the T-learner, which can also be regarded as a special case of the RA-learner \citep{curthNonparametricEstimationHeterogeneous2021}. The Doubly-Robust learner (DR-learner) employs pseudo-outcomes constructed from both the conditional outcomes $\hat{\mu}_t(x)$ and the propensity scores $\hat{\pi}(x)$ \citep{kennedyOptimalDoublyRobust2020}. DR estimators have a long history in causal inference and missing data imputation \citep{kangDemystifyingDoubleRobustness2007, bangDoublyRobustEstimation2005, robinsEstimationRegressionCoefficients1994, funkDoublyRobustEstimation2011} and offer the advantage of consistency as long as either the propensity score model or conditional outcome models are correctly specified. Finally, the R-learner \citep{nieQuasiOracleEstimationHeterogeneous2020a, fosterOrthogonalStatisticalLearning2023a} involves the formulation a specific loss function after fitting several nuisance functions that can be separately minimized and regularized to estimate the CATE, drawing inspiration from the Robinson decomposition \citep{robinsonRootNConsistentSemiparametricRegression1988}. There also exists CATE estimation methods that adapt specific ML models \citep{jacobCATEMeetsML2021a}. For instance, causal forests, introduced by \citet{atheyGeneralizedRandomForests2019, wagerEstimationInferenceHeterogeneous2018}, resemble the R-learner \citep{caronEstimatingIndividualTreatment2022, postFlexibleMachineLearning2022, oprescuOrthogonalRandomForest2019}.

A large body of literature analyzes the asymptotic and finite sample properties of different CATE learners \citep{salditt2023tutorial, curthNonparametricEstimationHeterogeneous2021, kunzelMetalearnersEstimatingHeterogeneous2019, caronEstimatingIndividualTreatment2022, kennedyOptimalDoublyRobust2020}. However, providing clear guidelines for determining the most suitable approach in real-world scenarios remains challenging. Which method will be most appropriate will generally depend on various factors, such as the level of confounding, the presence of high-dimensional covariates, the expected complexity of the CATE function compared to the individual outcomes and whether the treatment groups are strongly unbalanced. We recommend \citet{curth2023search} for a detailed discussion of the advantages and disadvantages of different CATE estimation strategies.

\section{Off-Policy Learning}
\label{appendix: off-policy learning}

Common approaches to construct an estimator for the policy value from observational data involve using weighting techniques, where propensity scores are estimated to re-balance the data, making it resemble data generated under the target policy \citep{kallusBalancedPolicyEvaluation2018, swaminathanCounterfactualRiskMinimization2015}. For instance, \citet{kitagawaWhoShouldBe2018} develop an algorithm that makes use of inverse propensity score weighting (IPW) to estimate $V(\pi)$ in a binary deterministic decision setting. Alternatively, some methods opt for direct estimation of the optimal treatment policy by fitting the outcome regression $\mathbb{E} [Y | X = x, T = t]$ and use the resulting estimates to optimize the policy value $\hat{V}$ \citep{qianPerformanceGuaranteesIndividualized2011, bennettEfficientPolicyLearning2020a}. Doubly Robust (DR) methods combine the IPW and direct approach by using an augmented IPW (AIPW) loss \citep{robinsEstimationRegressionCoefficients1994}. This requires estimating both the  propensity scores and the outcome regression model \citep{atheyPolicyLearningObservational2021, zhangEstimatingOptimalTreatment2012, dudikDoublyRobustPolicy2011a}. Several approaches have been proposed to relax the assumptions for causal identification, for example methods that address learning policies under unmeasured confounding \citep{bennettPolicyEvaluationLatent2019} or handle situations with limited overlap \citep{kallusMoreEfficientPolicy2021}.

\end{document}